%% file: for_arxiv.tex
\documentclass{article} 
\usepackage{iclr2027_conference,times}

\input{math_commands.tex}

\usepackage{hyperref}
\usepackage{url}
\usepackage{booktabs}
\usepackage{multirow}
\usepackage{graphicx}
\usepackage{amsmath}
\usepackage{subcaption}

\newcommand{\Vact}{V_{\mathrm{act}}}
\newcommand{\Vlat}{V_{\mathrm{latent}}}
\newcommand{\bon}{Agg-$N$}
\newcommand{\boe}{Agg-8}
\newcommand{\selEight}{Sel-8}

\usepackage{xcolor}

\title{Verify, Act, Distill: Turning Test-Time Compute into Test-Time Adaptation for Vision-Language-Action Models}

\author{
Kaneyoshi Hiratsuka \\
Kyoto University \\
\And
Satoshi Yamamori \\
Kyoto University \\
\And
Jun Morimoto \\
Kyoto University
}

\iclrfinalcopy

\begin{document}
\maketitle
\lhead{Preprint. Under review at ICLR 2027}

\begin{abstract}
Test-time compute (TTC), which scores multiple candidate actions at inference time, is a powerful way to improve the execution performance of vision-language-action (VLA) models, but existing methods keep the policy frozen, so the gain must be recomputed at every execution.
We propose \emph{Verify, Act, Distill}, a framework that \emph{consolidates the gains of TTC into the VLA itself}.
A fixed verifier scores candidates generated by a flow-matching VLA, the weighted aggregate is executed, and the observations and aggregated chunks from successful adaptation episodes are distilled directly into LoRA parameters with the standard flow-matching loss.
This reuses supervision that TTC has already produced during execution, without additional reinforcement learning or rollouts dedicated to teacher generation.
On the 40 LIBERO tasks with a SmolVLA 5k checkpoint, distilling from only 10 episodes per task raises the mean success rate of verifier-free single-candidate generation from 54.1\% to 70.6\%, retaining about 74\% of the improvement obtained by aggregating eight candidates.
Applying the same aggregation to the distilled policy further raises the success rate from 76.4\% to 82.3\%, with a particularly large gain on LIBERO-Long (44.0\% to 57.8\%), and the single-candidate policy also improves on LIBERO-plus, which contains distribution shifts.
These results show that using verification as a learning signal, rather than only as an execution-time selector, both retains the improvements of TTC and re-amplifies the effect of subsequent TTC.
\end{abstract}

\section{Introduction}
\label{sec:introduction}
Generative vision-language-action (VLA) policies can produce diverse action chunks from the same observation, and test-time compute (TTC) that scores such candidates with a learned evaluator exploits this diversity to elicit better actions \citep{nakamoto_steering_2025,kwok_robomonkey_2025}.
Because existing methods keep the base policy frozen, however, the good decisions made at inference time never reach the policy's parameters: even when the same task is executed repeatedly, the gain must be recomputed every time by generating and evaluating candidates from scratch.

This inefficiency hides an unexploited learning opportunity.
An execution process that evaluates candidates does not merely consume compute; it produces, by itself, high-quality supervision about which action to take for the current observation.
\emph{If this supervision were fed back into the VLA, the TTC gain could be carried over to single-candidate generation that needs no additional compute.}
Moreover, if the updated VLA then proposes better candidates, distillation would strengthen TTC itself rather than merely substitute for it.
This work examines this \textbf{retention and re-amplification of improvement}.
We target VLAs that still have room to adapt to their deployment environment, as when a pretrained model is introduced zero-shot or with little data into a new environment or task; as a controlled proxy we use a 5k-step checkpoint taken partway through training, and we assess the applicable range of the method with a 10k checkpoint and with LIBERO-plus, which contains distribution shifts.

The core of the proposed framework, \textbf{Verify, Act, Distill}, is to \emph{repurpose the candidate-aggregation result already computed for execution as teacher data for the policy}.
A fixed verifier scores the candidates generated by the VLA, the weighted aggregate is executed, and after a small number of adaptation episodes per task the observations and aggregated chunks of the successful trials update the LoRA parameters of the action expert with the standard flow-matching loss.
Adaptation requires neither a separate RL specialist nor value-gradient optimization of teacher actions.
Afterwards, the verifier can be removed for lightweight single-candidate inference or, when compute allows, reused for candidate aggregation once more.
Our method differs from existing approaches that distill improved actions in \textbf{where the supervision comes from}: PA-RL trains on actions selected and optimized by a critic \citep{mark2024policyagnostic}, PLD distills recovery actions from a residual RL specialist \citep{xiao_self-improving_2025}, and Q-Planning updates the evaluator from experience while keeping the policy fixed \citep{giridhar_beyond_2026}.
We instead recover supervision from routine candidate-aggregation execution under a fixed verifier and return the improvement to \emph{the VLA that proposes the candidates}, turning the sampling-and-verification procedure into an adaptation process that raises both single-candidate performance and the performance of subsequent TTC.

Our contributions are threefold.
\begin{enumerate}
\item \textbf{Direct distillation of verification-based execution decisions into the VLA:} an adaptation method that extracts the aggregated chunks already produced during TTC execution from successful trajectories and trains the VLA on them without additional RL optimization.
\item \textbf{Retention and re-amplification of TTC gains:} on the 40 LIBERO tasks with SmolVLA 5k, 10 adaptation episodes per task raise verifier-free single-candidate success from 54.1\% to 70.6\% (about 74\% of the eight-candidate gain), and re-applying aggregation to the distilled policy raises it from 76.4\% to 82.3\% (44.0\% to 57.8\% on Long). Single-candidate gains persist for a 10k checkpoint (83.2\% to 86.2\%) and under the distribution shifts of LIBERO-plus.
\item \textbf{Analysis of teacher selection and applicability:} we quantify the effect of the success filter and the teacher chunk, the behavior under distribution shift, and the dependence on the number of candidates and the adaptation budget.
\end{enumerate}
Retaining the improvement without a verifier is the primary goal; re-amplifying subsequent TTC is an additional benefit.

\section{Related Work}
\label{sec:related_work}

\paragraph{Inference-time improvement through candidate evaluation.}
Allocating additional inference-time compute to large models \citep{snell_scaling_2024}
has been widely applied to generative robot policies based on diffusion and flow matching \citep{chi_diffusion_2024,lipman_flow_2023}.
V-GPS re-ranks candidate actions with a value function \citep{nakamoto_steering_2025},
and RoboMonkey combines candidate generation from a VLA with a learned verifier \citep{kwok_robomonkey_2025}.
RoVer uses a robot reward model for candidate evaluation \citep{dai_rover_2025}, and TACO selects candidates based on an anti-exploration criterion \citep{yang_steering_2025}.
CoVer focuses on the consistency between the instruction and the action \citep{kwok_scaling_2026}, and VeriSpace evaluates actions based on spatial information \citep{zhao_verispace_2026}.
Other works use the VLA's own conditional likelihood as the selection criterion, as in MG-Select \citep{jang_verifier-free_2026},
or switch between selection and rejection depending on the quality of the candidate set, as in BOKBO \citep{singh_bokbo_2026}.
All of these focus on selecting or generating a good action to execute from a candidate set; our work differs in connecting that execution-time decision to the training of the candidate-generating policy itself.

\paragraph{Guidance and lookahead in the generative process.}
Inference-time techniques for improving generation include optimizing the initial noise \citep{nguyen_self-improving_2026}, guiding the flow with a Q-function \citep{yang_guided_2026},
combining latent-space RL with adaptive guidance \citep{tan_rl2-vla_2026}, and dynamically adjusting inference compute according to predictive uncertainty \citep{choi_scale_2026}.
Tree search has been introduced as in VLA-Reasoner \citep{guo_vla-reasoner_2026}, hierarchical policies have been combined with world models as in $\tau_0$-VLA \citep{cai__0-vla_2026},
and future-state prediction has been exploited through test-time scaling of World Action Models \citep{zhao_test-time_2026}.
SVA distills knowledge obtained from search into a Q-evaluator and evaluates the candidates of a frozen VLA at execution time \citep{xie2026look}.
In contrast, we distill the evaluation of the current candidate set directly into the VLA itself, improving the base performance of single-candidate generation.

\paragraph{Post-deployment adaptation and reinforcement learning.}
DSRL learns a policy that controls the input noise of a generative policy \citep{wagenmaker_steering_2025},
and TT-VLA and EVOLVE-VLA adapt VLAs using environment feedback from the deployment site \citep{liu_--fly_2026,bai_evolve-vla_2025}.
Reinforcement-learning approaches to further training VLAs include ConRFT \citep{chen_conrft_2025},
EXPO and EXPO-FT \citep{dong_expo_2026,dong_expo-ft_2026},
and VLA-RL, SimpleVLA-RL, and TGRPO \citep{lu_vla-rl_2025,li_simplevla-rl_2025,chen_tgrpo_2025}, which have been studied intensively.
The relationship between RL objectives and generalization \citep{liu_what_2026}, and frameworks that combine action-chunked PPO with imitation of self-collected successful trajectories \citep{wang_vla_2025}, have also been explored.
Our method is distinctive in using the aggregated chunks obtained in the deployment environment directly as teacher data in the standard flow-matching loss.

\paragraph{Reward and value models as learning signals.}
Vision-Language-Action-Critic \citep{zhai_vision-language-action-critic_2025}, Large Reward Models \citep{wu_large_2026},
and SARM2 \citep{chen_sarm2_2026} provide evaluation signals that support the further training of robot policies.
Q-VGM converts the action gradient of a critic into a supervision signal for the velocity field of a flow \citep{wang2026qvgm}.
The verifier used in this work is trained in advance to rank candidates by their distance to the recorded actions of official demonstrations and successful rollouts.
At adaptation time, the aggregated action based on this ranking is adopted as the teacher, which completely separates the training of the evaluator from the update of the policy.

\paragraph{Reusing improved actions and policy distillation.}
Expert Iteration is a representative framework that trains the policy on search results and reuses the updated policy in subsequent search \citep{anthony2017thinking}.
PA-RL connects Q-function-based candidate selection and local optimization to supervised policy updates, and has been applied to OpenVLA and other models \citep{mark2024policyagnostic}.
PLD distills corrective actions generated by a residual RL specialist into the VLA \citep{xiao_self-improving_2025}.
Q-Planning combines weighted execution of action candidates with continual updates of the Q-function \citep{giridhar_beyond_2026}.
By using the execution-time candidate-aggregation result under a fixed verifier as the teacher, our work makes the process that generates teacher data identical to ordinary candidate evaluation and execution.

Data-generation approaches for self-improvement include imagined trajectories from world models \citep{yang_rise_2026}, reflection on and reuse of execution experience \citep{li_reflection-based_2026},
and collection of successful rollouts for novel skills \citep{wang_insight_2026}.
Refined Policy Distillation distills VLA knowledge into a small RL expert \citep{julg_refined_2025},
and DEAS improves long-horizon decision making and action prediction through value learning over action sequences \citep{kim_deas_2025}.

\section{Verify, Act, Distill}
\label{sec:method}
\begin{figure}[t]
\centering
\includegraphics[width=0.95\linewidth]{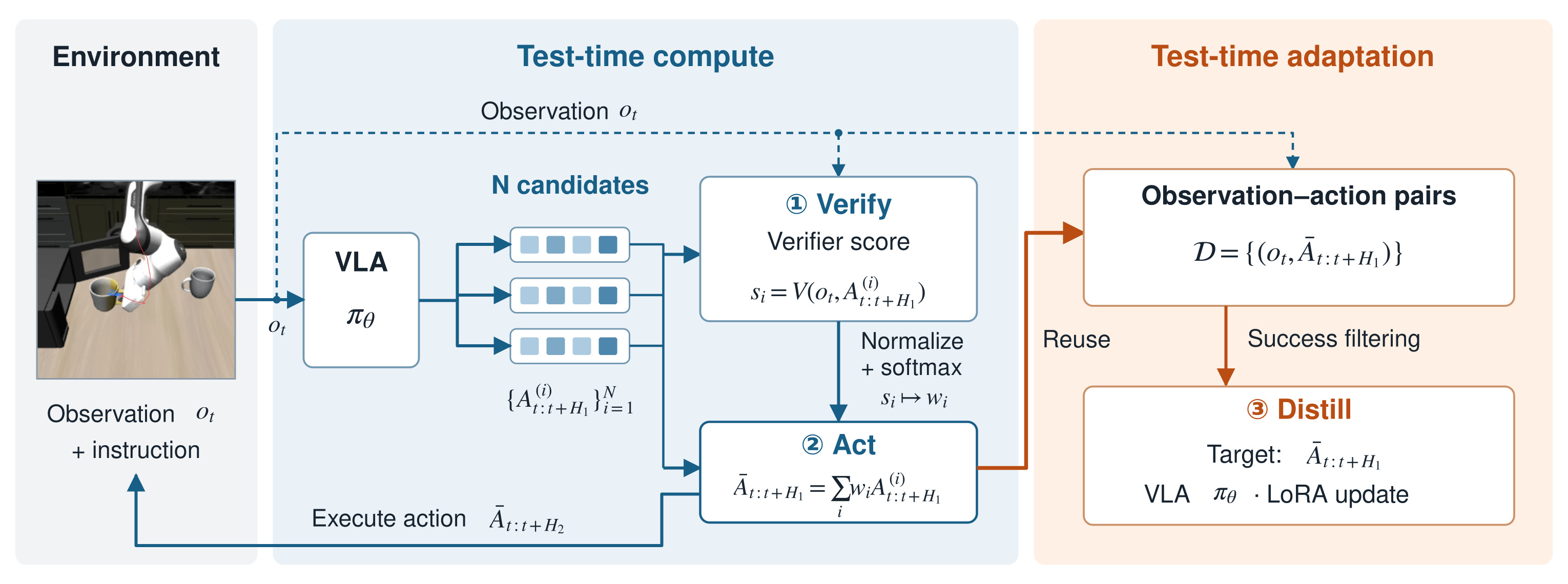}
\caption{Overview of Verify, Act, Distill. Candidate chunks are scored and aggregated by the verifier and the leading $H_2=10$ steps are executed.
Observations and the aggregated chunks of length $H_1=50$ from successful episodes are stored in a buffer and used to update the LoRA parameters.
The verifier evaluates only the first $H_2$ steps of each candidate, whereas the teacher data include the entire planned chunk.}
\label{fig:overview}
\end{figure}

Our method converts successful rollouts of a candidate-aggregation policy into teacher data for the same VLA; the policy and verifier are fixed during adaptation, and only the LoRA parameters are updated after episode collection (Figure~\ref{fig:overview}).

\subsection{Problem Setting}
\label{sec:prelim}
The observation $o_t$ consists of two camera images, an 8-dimensional proprioceptive state, and a language instruction.
A pretrained VLA $\pi_{\theta_0}$ maps an initial noise $z$ and $o_t$ to a normalized action chunk $A=\pi_{\theta_0}(o_t,z)\in\mathbb{R}^{H_1\times d}$; we use SmolVLA \citep{shukor_smolvla_2025} with $H_1=50$ and $d=32$.
The leading 7 dimensions are applied to the environment through $g$, which undoes the normalization and applies the environment-specific coordinate transformation, and only the first $H_2=10$ steps $g(A_{0:H_2})$ (half-open interval) are executed at each replanning point; $H_1$ is the generation horizon and $H_2$ the execution horizon.

For each task we collect $K$ adaptation episodes, update the parameters once, and evaluate under new reset seeds. The LoRA parameters and buffer are initialized per task, no updates occur during evaluation, and nothing is transferred across tasks.
During adaptation the only available information is the observations, the generated candidates, the verifier scores, and the episode-level success outcome; offline pretraining uses newly collected successful rollouts for the policy and, additionally, the official LIBERO demonstrations for the verifier.
To decide the executed action from candidates we use a weighted average (\emph{candidate aggregation}) rather than the argmax; ``Best-of-$N$'' in Figure~\ref{fig:overview} denotes this operation, and for $N=1$ a single candidate is executed without invoking the verifier.

\subsection{Verify and Act: Candidate Aggregation with a Fixed Evaluator}
\label{sec:ttc}
\paragraph{Candidate generation.}
At replanning point $b$ we draw $N$ candidates ($N=8$ by default) from independent noise: the first from $z^{(1)}\sim\mathcal{N}(0,I)$ and the rest from $z^{(i)}\sim\mathcal{N}(0,\alpha^2I)$ with $\alpha=1.2$, so that one action from the base policy's own distribution is always in the set (Appendix~\ref{app:alpha} compares $\alpha$).

\paragraph{Verifier.}
Following RoboMonkey \citep{kwok_robomonkey_2025}, the action evaluator $\Vact$ maps the observation features and the environment actions of the first $H_2$ steps of a candidate to a scalar $s_i=\Vact(o_b,g(A^{(i)}_{0:H_2}))$.
Observation features are the frozen VLM connector outputs (two views, 64 tokens each, 960 dimensions) of a checkpoint trained for 20k steps, so even the 5k policy is evaluated with a stable representation; the evaluator is a 6-layer Transformer with hidden size 512, and candidate data are created and a separate verifier trained for each of the 5k, 10k, and 20k policies (Appendix~\ref{app:verifier}).
The training data consist of eight candidates generated at each state of the official LIBERO demonstrations (all four suites) and of newly collected successful rollouts; letting $d_i$ be the environment-space RMSE between the first $H_2$ steps of candidate $i$ and the 10 recorded action steps of the source trajectory at that state (the demonstration actions for demonstration states and the rollout's own actions for rollout states), we minimize an all-pairs Bradley--Terry loss \citep{bradley_terry} preferring $i$ over $j$ whenever $d_i<d_j$.
The verifier thus learns to score candidates by their closeness to the recorded actions of successful trajectories. Training runs for 60k steps with batch size 16 and learning rate $10^{-4}$ (Appendix~\ref{app:verifier}).

\paragraph{Weighted execution.}
Scores are standardized and converted to softmax weights with temperature $\lambda=0.3$:
\begin{equation}
 w_i=\frac{\exp(u_i)}{\sum_{j=1}^{N}\exp(u_j)},\qquad
 u_i=\frac{s_i-\max_j s_j}{\lambda(\sigma_s+\delta)},\qquad
 \bar A_b=\sum_{i=1}^{N}w_i A^{(i)},
 \label{eq:softmax}
\end{equation}
where $\sigma_s$ is the score standard deviation within the set and $\delta=10^{-8}$ ($w_1=1$ for $N=1$).
Aggregation is performed in the normalized space and the first $H_2$ steps of $\bar A_b$ are executed.
We denote weighted aggregation over $N$ candidates by \bon\ (\boe\ for $N=8$) and argmax selection among eight by \selEight; Appendix~\ref{app:softmax} compares the two execution rules.

\subsection{Distill: Supervised Update from Successful Rollouts}
\label{sec:tta}
In each adaptation episode $e$, the pair $(o_b,\bar A_b)$ at every replanning point is stored temporarily, and with the success label $y_e\in\{0,1\}$ the distillation buffer is
\begin{equation}
 \mathcal D=\bigcup_{e:y_e=1}\{(o_b,\bar A_b):b\text{ is a replanning point of episode }e\},
 \label{eq:buffer}
\end{equation}
with $K=10$ episodes by default. If no episode succeeds, the update is skipped and the base policy is kept.
Only the LoRA parameters $\phi$ of the action expert are updated; $\theta_0$ and the verifier stay fixed.
With $A_\tau=(1-\tau)\bar A+\tau\epsilon$, $\tau\sim U(0,1)$, $\epsilon\sim\mathcal N(0,I)$, the LoRA minimizes the flow-matching loss \citep{lipman_flow_2023,hu_lora_2021}
\begin{equation}
 \mathcal L(\phi)=
 \mathbb E_{(o,\bar A)\sim\mathcal D,\tau,\epsilon}
 \left[\left\|v_{\theta_0,\phi}(A_\tau,\tau\mid o)-(\epsilon-\bar A)\right\|_F^2\right],
 \label{eq:fm}
\end{equation}
which is formulated in the data-to-noise direction and integrated in the reverse direction at inference.
The teacher is the full $H_1$-step aggregated chunk: although only the first $H_2$ steps were executed, imitating the whole plan keeps the same input/output dimensions and objective as SmolVLA's pretraining \citep{shukor_smolvla_2025}, and because the weights $w_i$ are shared across time steps, the mixture chosen by evaluating the leading segment yields one coherent plan rather than a patchwork of candidates (Appendix~\ref{app:h2only} compares a teacher masked to the executed $H_2$ steps).
The LoRA has rank 8, scaling 16, and 0.37M trainable parameters, updated with Adam (learning rate $10^{-4}$, batch 8) for 200 steps (Appendix~\ref{app:distill-impl}).

\subsection{Usage after Adaptation and Cost}
\label{sec:cost}
We evaluate the \textbf{distilled policy (single candidate)}, which runs with the same number of candidates and inference steps as the base policy and needs neither the verifier nor its observation encoder, and the \textbf{distilled policy + \bon}, which reuses the same fixed verifier to test the synergy between the update and candidate quality.
When the $K$ rollouts are collected during ordinary TTC execution, the only extra cost is storing the executed data and 200 LoRA steps (measured latency on real hardware remains to be examined).

\section{Experimental Results}
\label{sec:results}
We examine (1) whether the improvement is retained after removing the verifier, (2) whether re-applying aggregation to the distilled policy yields further gains, and (3) how the success filter and evaluation conditions affect the outcome.

\subsection{Setup and Protocol}
\label{sec:setup}
We use the four LIBERO suites \citep{liu_libero_2023} (Spatial, Object, Goal, Long; 10 tasks each).
SmolVLA is trained on the 5,225 successful episodes among 10,000 rollouts collected with $\pi_{0.5}$ \citep{intelligence__05_2025}; the main experiments use the 5k checkpoint as a proxy for an insufficiently adapted policy, and we also evaluate the 10k checkpoint and, with it, LIBERO-plus \citep{fei_libero-plus_2025} under distribution shift.
On clean LIBERO, each policy is adapted with reset seeds 0--9 and evaluated on held-out seeds 10--59 (50 episodes per task), which are also used for the unadapted Base and \boe\ (data, training, compute, episode limits, and success criteria are given in Appendices~\ref{app:impl} and~\ref{app:protocol}).
We report suite means, the four-suite macro-average, per-task paired differences (mean $\pm$ SEM over tasks; Appendix~\ref{app:protocol}), and the number of tasks that improved / were unchanged / degraded (positive/zero/negative).
On LIBERO-plus we use 105 tasks per suite (15 from each of seven perturbation categories) with adaptation seeds 0--9 and evaluation seeds 10--19; due to compute constraints, the distilled policy is evaluated on the first 35 tasks per suite (140 tasks), and Base and \boe\ are aggregated on the same subset, which is not guaranteed to be balanced across categories.

\paragraph{Compared methods and notation.}
Base is the unadapted policy; Self distills Base's own successful rollouts without a verifier (Appendix~\ref{app:self-imitation}); Distill is the proposed method; and N-GRPO is noise-space GRPO, which keeps the VLA and verifier fixed and learns a perturbation distribution over the input noise by RL with the verifier ranking as reward, updating only from successful episodes (Appendices~\ref{app:grpo} and~\ref{app:baselines}).
Names without a suffix denote single-candidate generation; $+$\boe\ and $+$\selEight\ denote evaluation with weighted aggregation and argmax selection over eight candidates, respectively.

\subsection{Main Results: Retention and Synergy with Candidate Aggregation}
\label{sec:main}

\begin{figure}[t]
\begin{minipage}[c]{0.5\linewidth}
\centering
\captionof{table}{Success rates (\%) on LIBERO (clean), SmolVLA 5k, evaluation seeds 10--59 (50 episodes per task, 500 per suite).
Self, Distill, and N-GRPO use 10 adaptation episodes; \boe\ is weighted aggregation with $\Vact$ ($\lambda=0.3$).}
\label{tab:main5k}
\footnotesize
\setlength{\tabcolsep}{2pt}
\begin{tabular}{lccccc}
\toprule
Method & Spatial & Object & Goal & Long & Avg. \\
\midrule
Base               & 57.6 & 80.4 & 61.0 & 17.4 & 54.1 \\
\boe                     & 80.6 & 94.4 & 86.8 & 44.0 & 76.4 \\
\midrule
N-GRPO            & 64.8 & 79.2 & 68.2 & 20.0 & 58.1 \\
N-GRPO + \boe           & 84.4 & 95.0 & 86.6 & 50.6 & 79.2 \\
\midrule
Self & 74.4 & 89.8 & 63.2 & 27.6 & 63.7 \\
Self + \boe & 86.8 & \textbf{98.6} & 82.6 & 55.2 & 80.8 \\
\midrule
Distill        & 81.8 & 88.0 & 79.4 & 33.2 & 70.6 \\
Distill + \boe       & \textbf{88.2} & 95.4 & \textbf{87.8} & \textbf{57.8} & \textbf{82.3} \\
\bottomrule
\end{tabular}
\end{minipage}\hfill
\begin{minipage}[c]{0.48\linewidth}
\centering
\includegraphics[width=\linewidth]{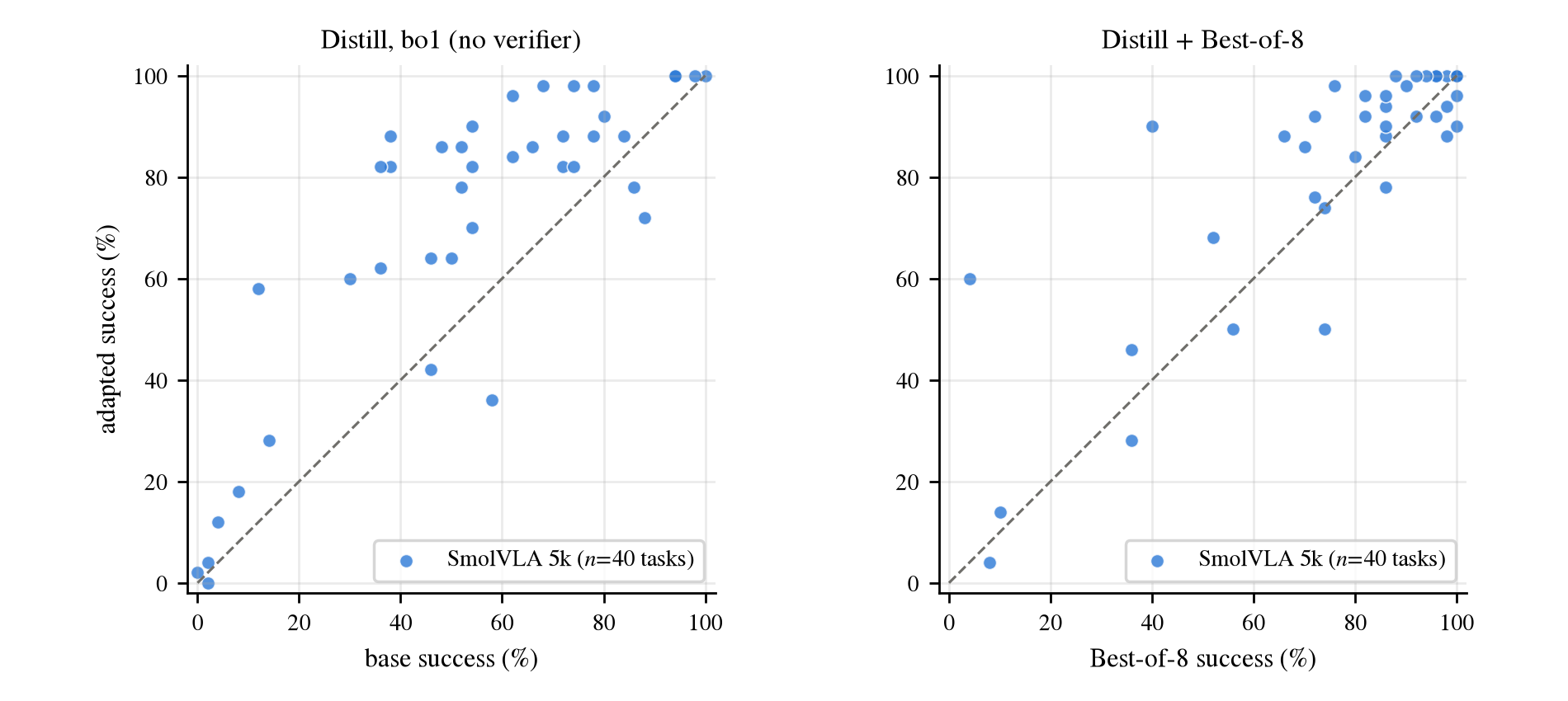}
\caption{Per-task success rates (SmolVLA 5k, 40 tasks, 50 episodes per point, seeds 10--59).
Left: Base vs.\ distilled policy with a single candidate (no verifier); right: \boe\ vs.\ distilled policy + \boe. Points above the diagonal indicate improvement (left 34/1/5, right 24/6/10).}
\label{fig:scatter}
\end{minipage}
\end{figure}

\paragraph{About 74\% of the TTC gain is retained in a verifier-free policy.}
As Table~\ref{tab:main5k} shows, aggregating eight candidates (\boe) raises the mean success rate substantially, from 54.1\% to 76.4\%.
The policy distilled from the successful cases of this aggregated execution (Distill) then reaches 70.6\% with verifier-free single-candidate generation.
The per-task paired difference against Base averages $+16.5\pm2.7$ points; 34 of the 40 tasks improve, 1 is unchanged, and only 5 degrade (positive/zero/negative: 34/1/5).
The fraction of the TTC gain that is retained, $(70.6-54.1)/(76.4-54.1)$, is about 74\%.

\paragraph{Re-amplification of subsequent TTC through distillation.}
Re-applying candidate aggregation to the distilled policy (``Distill + \boe'') yields a mean success rate of 82.3\%, a paired difference of $+5.8\pm2.3$ points over the original \boe.
The gain is especially large on the Long suite, which demands long-horizon planning: from 44.0\% to 57.8\%.
At the task level, however, the breakdown is 24 improved, 6 tied, and 10 degraded, so some tasks do get worse (Figure~\ref{fig:scatter}).
These results show that transferring the TTC gain into single-candidate generation and raising the ceiling of subsequent candidate aggregation can be achieved simultaneously.

\paragraph{Gains beyond self-imitation of successes.}
As a control, we evaluated ``Self,'' which distills only the base policy's own successful rollouts without using the verifier (Table~\ref{tab:main5k}; implementation details in Appendix~\ref{app:self-imitation}).
Self also improves single-candidate success from 54.1\% to 63.7\%, but Distill, which uses the verifier-aggregated chunks, reaches 70.6\%, a further 6.9 points higher.
By suite, Distill exceeds Self by 7.4 points on Spatial, 16.2 points on Goal, and 5.6 points on Long; only on Object is Self higher, by 1.8 points.
With candidate aggregation as well, Distill records 82.3\% against Self's 80.8\%.
Although simple self-imitation of successes gives some improvement, routing the teacher through the verifier's higher-quality aggregation consolidates the gain into the single-candidate policy more strongly, especially on suites such as Goal.

During the 10 adaptation episodes, the trial success rate was 53.5\% for Self and 73.8\% for Distill, and the mean number of transitions accumulated in the distillation buffer was 84 boundaries for Self versus 111 for Distill.
For Self, 3 of the 40 tasks yielded no successful example during adaptation, and the base policy was used for evaluation as is.
The observed additional improvement therefore reflects the combined effect of the whole adaptation process: more successful data collected and higher-quality aggregated chunks.

\paragraph{Comparison with noise-space GRPO.}
Under the same adaptation/evaluation seeds and budget $K=10$, we evaluated noise-space GRPO, which uses the verifier ranking as the reward signal.
It reaches only 58.1\% with single-candidate generation and 79.2\% with candidate aggregation, consistently below the proposed distillation (70.6\% / 82.3\%).
Under a limited adaptation-episode budget, distilling the aggregated action chunks as direct supervision for the VLA is far more sample-efficient than using the verifier ranking as an RL reward for a noise policy.
Implementation details of the comparison are given in Appendix~\ref{app:baselines}, and additional evaluations with argmax execution and alternative advantage computations in Appendix~\ref{app:grpo-eval}.

\subsection{Retention of Improvement at the 10k Checkpoint}
\label{sec:strong}
For the further-trained 10k checkpoint, the distilled single-candidate policy records 86.2\% against Base's 83.2\%
(Table~\ref{tab:main10k}; paired difference $+3.0\pm1.4$ points).
The primary goal of our method, obtaining the improvement without additional candidate generation or verifier inference, is thus also achieved for a more thoroughly trained policy.
The retention effect is therefore not specific to the 5k condition, where Base's success rate was as low as 54.1\%.

On the other hand, combining the distilled policy with candidate aggregation gives 89.8\%, which does not exceed the 90.4\% of the original candidate aggregation.
At the 10k checkpoint, then, the effect of distillation appears mainly as improved single-candidate generation, and no re-amplification of TTC is observed.
One possible factor is that a fixed verifier trained on distance to the recorded actions of successful trajectories has difficulty discriminating the small differences among the high-quality candidates generated by a sufficiently improved policy.
Indeed, the verifier's validation pairwise accuracy drops from 0.82 for the 5k policy to 0.73 for the 10k policy. Because the candidate distribution itself also changes, however, this number alone does not allow us to conclude that reduced discriminative ability is the main cause.
Other factors, such as reduced candidate diversity or a shift in the post-distillation generative distribution, may also contribute.
Overall, retention of improvement is reliably obtained at 10k as well, but the additional amplification from candidate aggregation is more pronounced at 5k (Appendix~\ref{app:tab1-10k}).
Distillation and aggregation at the 20k checkpoint, comparisons with other adaptation methods, and the per-checkpoint reproducibility of the correction direction under re-sampled candidate sets are given in Appendices~\ref{app:strength}, \ref{app:20k}, and~\ref{app:amort}.

\subsection{Ablation: Success Filter and Teacher Chunk}
\label{sec:ablation}
Distilling from all adaptation episodes without the success filter gives 64.4\% / 78.8\% (single candidate / aggregation), clearly below 70.6\% / 82.3\%; the paired differences in favor of the filter are $+6.2\pm1.6$ and $+3.5\pm1.7$ points (Appendix~\ref{app:distill-abl}).
Using the highest-scoring candidate instead of the aggregated chunk as the teacher gives 67.8\% / 82.5\% ($-2.8\pm2.0$ and $+0.2\pm1.8$ points relative to the aggregated teacher); we keep the aggregated chunk as the default because it matches the plan that was actually generated and executed during adaptation.

\subsection{Distribution Shift and Scarce Successes}
\label{sec:plus}

\begin{table}[t]
\caption{Success rates (\%) on LIBERO-plus, SmolVLA 10k, 105 tasks per suite, seeds 10--19 ($n{=}1050$ per suite).
The lower block uses the first 35 tasks per suite (140 tasks, $n{=}350$), on which Base and \boe\ are re-aggregated for comparison.}
\label{tab:plus10k}
\begin{center}
\small
\begin{tabular}{lccccc}
\toprule
Method (10k, LIBERO-plus) & Spatial & Object & Goal & Long & Avg. \\
\midrule
Base                     & 35.5 & 18.8 & 13.0 & 16.2 & 20.9 \\
\boe\ ($\Vact$)           & 41.5 & 24.9 & 14.3 & 21.7 & 25.6 \\
\midrule
\multicolumn{6}{l}{\emph{First 35 tasks per suite (common subset)}} \\
Base                     & 36.6 & 15.1 & 14.9 & 16.0 & 20.7 \\
\boe\ ($\Vact$)           & 40.6 & \textbf{17.7} & \textbf{17.1} & \textbf{23.4} & 24.7 \\
Distill      & 41.1 & \textbf{17.7} & 16.6 & 22.6 & 24.5 \\
Distill + \boe     & \textbf{46.3} & 16.9 & 15.7 & 22.6 & \textbf{25.4} \\
\bottomrule
\end{tabular}
\end{center}
\end{table}

Table~\ref{tab:plus10k} reports the results on LIBERO-plus. Over all 105 tasks per suite, the 10k policy averages 20.9\% for Base and 25.6\% for \boe.
On the common subset of 35 tasks per suite used to evaluate the distilled policy, Base reaches 20.7\%, \boe 24.7\%, the distilled policy (single candidate) 24.5\%, and the distilled policy + \boe 25.4\%.
The paired difference of distilled single-candidate generation over Base is $+3.9\pm1.4$ points (positive/zero/negative: 43/74/23),
and the difference of the aggregated variant over the original \boe is $+0.6\pm1.2$ points (positive/zero/negative: 34/76/30).
Even under distribution shift with unseen perturbations, single-candidate generation improves, but the additional synergy with candidate aggregation is limited.

Moreover, in 66 of the 140 tasks no success was obtained during the adaptation trials, so no update was performed and the base policy was retained.
That is, teacher data for distillation were recovered only from the remaining 74 tasks that succeeded at least once during adaptation.
An evaluation in which the verifier input is replaced by latent variables predicted by a world model is given in Appendix~\ref{app:vlat}, and a comparison of feature spaces in Appendix~\ref{app:encoder}.

On the same subset, applying the distillation variant without the success filter yields a single-candidate paired difference over Base of only $+0.7\pm1.1$ points, so the improvement essentially vanishes.

\subsection{Number of Candidates and Adaptation Budget}
\label{sec:budget}
\begin{figure}[t]
\centering
\begin{subfigure}[t]{0.56\linewidth}
\centering
\includegraphics[width=\linewidth]{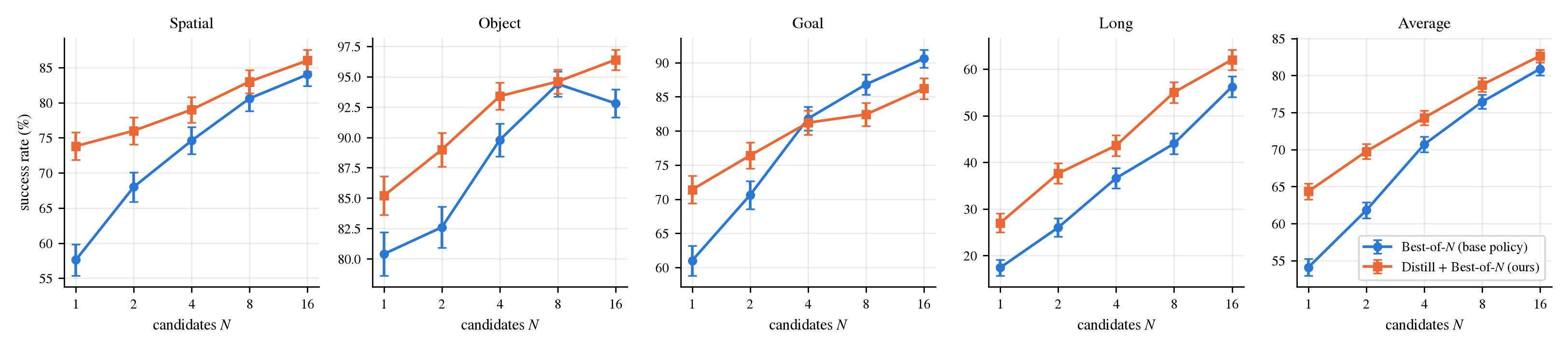}
\caption{Success rate vs.\ $N$ for \bon\ (base policy) and distilled policy + \bon\ (one LoRA distilled with $K{=}10$); SmolVLA 5k, $n{=}500$ per point, seeds 10--59, binomial SE.}
\label{fig:nsweep}
\end{subfigure}\hfill
\begin{subfigure}[t]{0.42\linewidth}
\centering
\includegraphics[width=\linewidth]{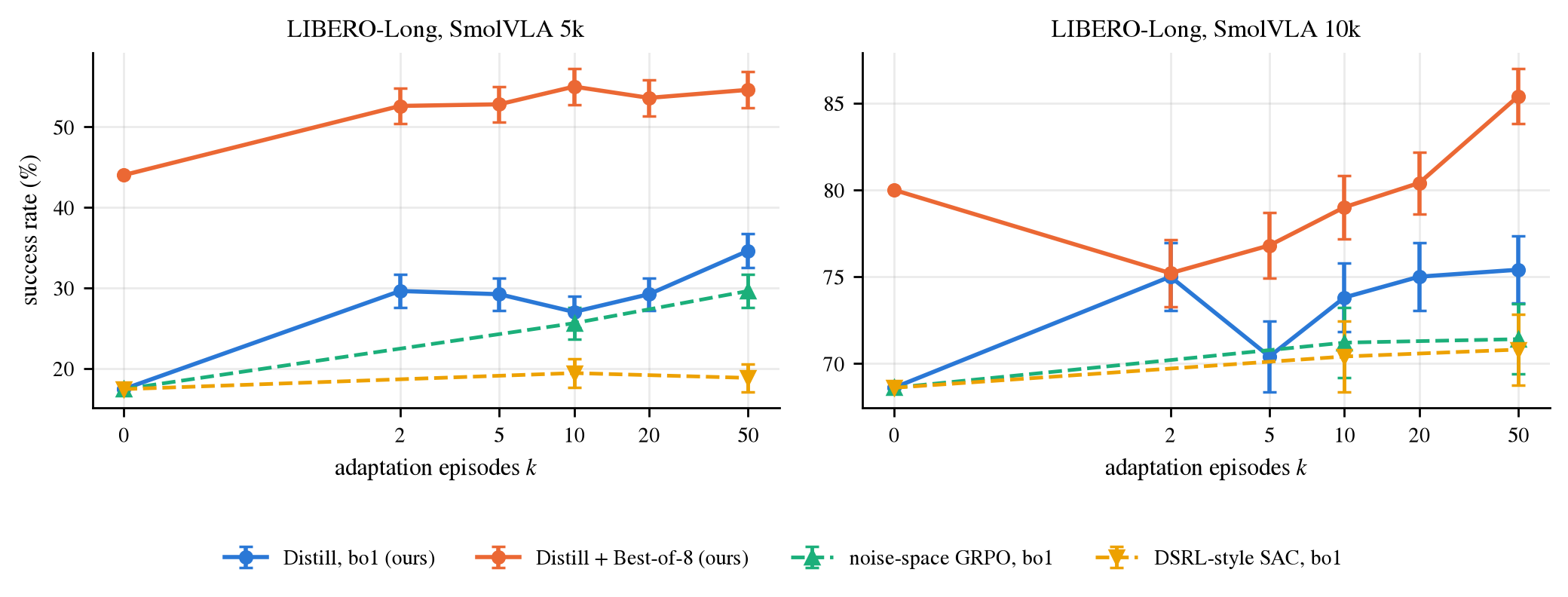}
\caption{Success rate vs.\ adaptation budget $K$ on LIBERO-Long ($n{=}500$ per point; $K{=}0$ is Base and \boe); the evaluation seed set differs per $K$.}
\label{fig:budget}
\end{subfigure}
\caption{Dependence on the number of candidates and the adaptation budget. Both use the variant without the success filter; numbers are in Tables~\ref{tab:nsweep} and~\ref{tab:budget}.}
\label{fig:sweeps}
\end{figure}
To characterize the dependence on the number of candidates $N$ and the adaptation-episode budget $K$, we ran additional experiments with the variant that does not use the success filter (Appendix~\ref{app:sweeps}).
As shown in Figure~\ref{fig:nsweep}, for every number of candidates evaluated, $N\in\{1,2,4,8,16\}$, the four-suite mean success rate with the distilled policy consistently exceeds that with the original base policy.
The improvement from distillation is larger in the regime of few candidates, and on the Long suite a 5.8-point improvement is maintained even with as many as $N=16$ candidates.
This indicates that the benefit of distillation is not overfitted to the number of candidates (eight) used during training.
By suite, however, Goal shows saturation or a decline for $N\ge4$.

In the adaptation-budget experiment (Figure~\ref{fig:budget}), on the Long suite with the 5k checkpoint the single-candidate success rate reaches 29.6\% with only $K=2$ and 34.6\% at $K=50$.
Under the same budget $K=10$, the proposed distilled policy achieves 27.0\%, noise-space GRPO 25.6\%, and DSRL-style SAC 19.4\%.
These trends corroborate the very high data efficiency of our approach, which recovers supervision from a handful of rollouts and adapts the policy.
The evaluation seeds here are $K,\ldots,K+49$, and the performance after collection at each budget is measured separately.

\subsection{Qualitative Example}
\label{sec:qual}
\begin{figure}[t]
\begin{center}
\includegraphics[width=0.95\linewidth]{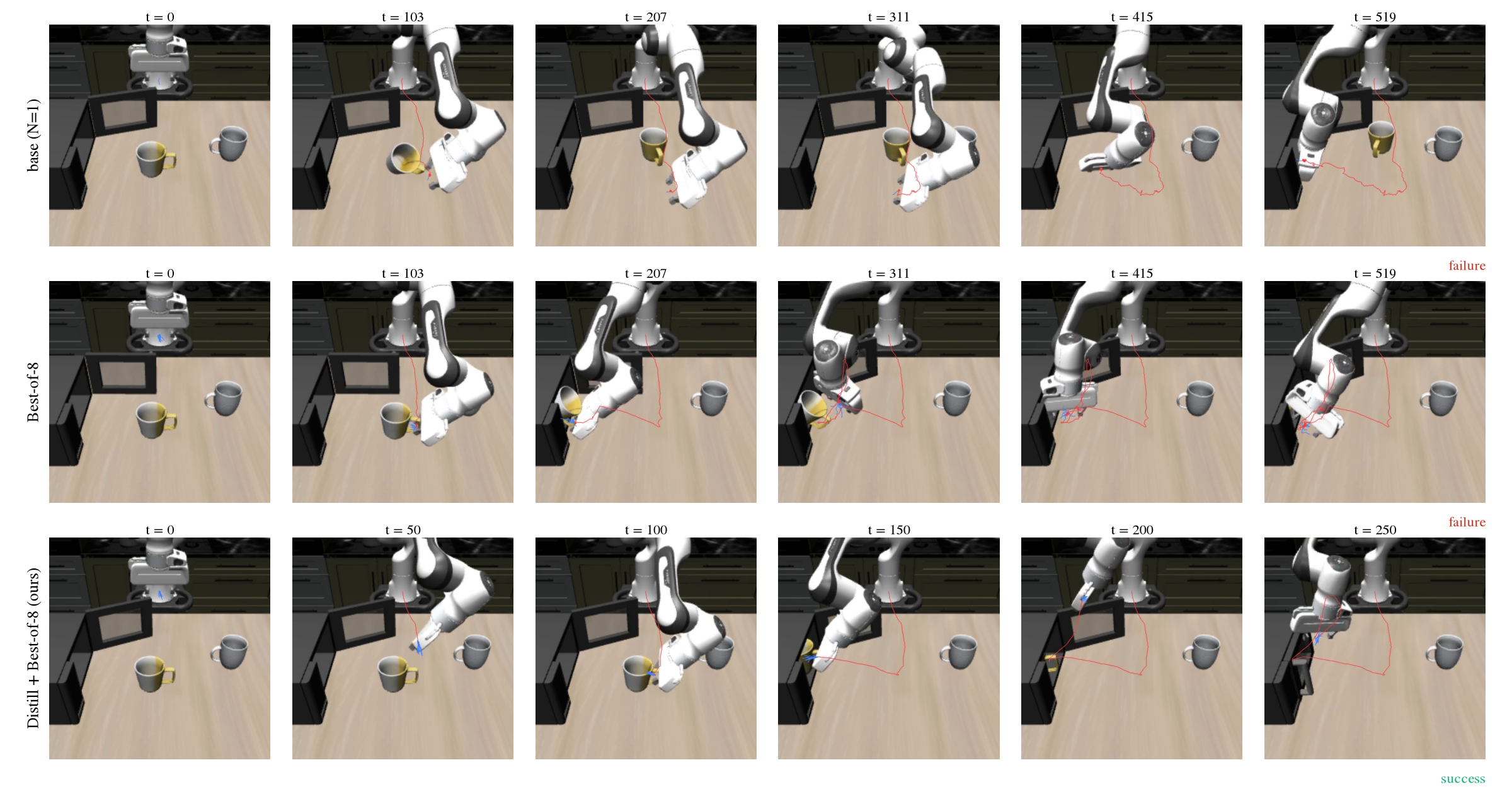}
\end{center}
\caption{Rollouts on LIBERO-Long task 9 (SmolVLA 5k) from the same initial state (seed 12) for Base ($N{=}1$), \boe, and distilled policy + \boe; thin lines show each candidate's predicted trajectory at each chunk boundary and the thick line the executed weighted average, and rows show the progression of time.
The seed is the first one, searching upward from seed 10, in which Base fails and the distilled policy + \boe\ succeeds (\boe\ alone also fails here).}
\label{fig:anim}
\end{figure}
Figure~\ref{fig:anim} shows a case in which Base and ordinary aggregation (\boe) both fail from the same initial state while only the distilled policy + \boe\ succeeds: where the former break down during grasping or manipulation, the higher-quality candidates of the distilled policy lead the rollout stably to completion.

\section{Conclusion and Limitations}
\label{sec:conclusion}
Verify, Act, Distill extends inference-time verification in VLAs from a one-off action selector into a learning signal for the policy.
By extracting aggregated actions from successful candidate-aggregation rollouts and writing them back into the VLA with a single LoRA update, it retains about 74\% of the TTC gain in verifier-free single-candidate generation on LIBERO and further improves subsequent candidate aggregation.
TTC need not be compute consumed anew at every inference; it can generate the teacher data that improves the very policy responsible for future inference.

\paragraph{Limitations.}
\label{sec:limitations}
Our method rests on two prerequisites: (1) the verifier must discriminate action candidates that lead to improvement, and (2) at least some adaptation episodes must succeed.
The first ties the teacher to a fixed verifier trained on distances to recorded actions, so the ``no additional data'' property holds only for the adaptation phase, and distillation may pull the policy toward the style of those actions and reduce candidate diversity, which is consistent with the limited re-amplification at 10k (\S\ref{sec:strong}).
The second binds under severe distribution shift: on LIBERO-plus, 66 of the 140 tasks yielded no success within 10 episodes and could not be updated (\S\ref{sec:plus}), and in the real world the success signal would come from a detector whose errors propagate directly into the distillation buffer.
The update is also unconditional, with no validation step or fallback to the base policy, and 5 of the 40 tasks degrade with the distilled single-candidate policy and 10 when aggregation is re-applied (Figure~\ref{fig:scatter}).
Finally, all results are in simulation with a single VLA architecture, a single training seed, and per-task LoRA parameters that are not transferred across tasks, and our cost argument counts candidates and evaluator calls rather than measured latency (\S\ref{sec:cost}).
Addressing these points, in particular safeguards against regressions and real-robot validation, is left to future work.

\section*{AI use statement}
We used generative AI tools (Claude and ChatGPT/Codex) to assist with implementing the experimental pipeline, aggregating results, interpreting experimental findings, and providing feedback on the methodology and experiments.
Additionally, we used these tools to create plotting and analysis scripts, identify and summarize relevant literature, draft and edit the manuscript, and organize its structure.
The rollout data used for training were generated by the pretrained robot policy described in Appendix~\ref{app:data}.
The authors checked AI-assisted experimental code and numerical results, and reviewed the AI-assisted text and interpretations against the experimental records and cited sources.
We take responsibility for the final content of this work, including text, claims, and artifacts produced with the aid of generative AI.

\section*{Reproducibility statement}
The benchmark environment, policy checkpoints, seeds used for adaptation and evaluation, and the procedure for aggregating success rates are specified in \S\ref{sec:setup}.
The candidate generation and aggregation algorithm and the distillation objective are detailed in \S\ref{sec:ttc} and \S\ref{sec:tta}.
The collection procedure for the pretraining data and the breakdown of its uses are given in Appendix~\ref{app:data} and Table~\ref{tab:training-data}, the verifier's network configuration and training parameters in Appendix~\ref{app:verifier},
and the implementation details of the compared methods and the LoRA update hyperparameters in Appendices~\ref{app:baselines} and~\ref{app:distill-impl}.
All experiments were run on NVIDIA RTX PRO 6000 (96 GB) and RTX 6000 Ada (48 GB) GPUs (Appendix~\ref{app:compute}).

\section*{Acknowledgments}
This work was supported by the Japan Science and Technology Agency (JST) through the ASPIRE program (JPMJAP2503), ACT-X program (JPMJAX25CR), and Moonshot Research and Development Program (JPMJMS263I).

\bibliography{iclr2027_conference}
\bibliographystyle{iclr2027_conference}
\appendix

\section{Implementation Details}
\label{app:impl}

\subsection{Data Collection and Policy Training}
\label{app:data}
The policy training data consist of 10,000 episodes collected by running a pretrained $\pi_{0.5}$ checkpoint on every task of the four LIBERO suites (250 episodes per task = 100 in the standard environment + 150 in a failure-injection environment). Failure injection deliberately adds an offset to the 6 pose dimensions of the environment action that increases linearly within a chunk (ee\_ramp, amplitude $r \in \{0.6, \dots, 1.6\}$, tuned per suite so that the success rate falls between 0.14 and 0.76). Of the collected data, only the 5,225 successful episodes that completed their task (808,143 frames) were used for policy training
(the official LIBERO demonstrations are not used for behavior cloning (BC) of the policy; they are used only for training the verifier and the world model described below).
Preliminary experiments showed that mixing the official demonstrations with the collected rollouts for BC degraded policy performance, so policy pretraining uses only the 5,225 newly collected successful episodes.

SmolVLA is based on lerobot's \texttt{smolvla\_base}; all parameters, including the vision encoder, were trained for 20k steps (batch size 32, lerobot's default learning rate, random seed 1000), and checkpoints were saved at 5k, 10k, 15k, and 20k steps.

Table~\ref{tab:training-data} summarizes the use of training data for each trained component.

\begin{table}[h]
\centering
\small
\caption{Breakdown of offline training data by use. This is distinct from the success-filtered distillation buffer used in the adaptation phase.}
\label{tab:training-data}
\begin{tabular}{p{0.22\linewidth}p{0.69\linewidth}}
\toprule
Trained component & Data and supervision \\
\midrule
SmolVLA (BC, 20k steps) & Only the 5,225 successful episodes among the newly collected rollouts. Official LIBERO demonstrations not used. \\
Verifier & Eight candidates generated at states from both the official LIBERO demonstrations (all four suites) and the newly collected successful rollouts. Supervision is the RMSE to the $H_2=10$ recorded action steps at each state. The 10k candidate dataset comprises 104,294 states. \\
World model (appendix) & Official LIBERO demonstrations and newly collected rollouts. For rollouts, both successful and failed data are used. \\
\bottomrule
\end{tabular}
\end{table}

\subsection{Hyperparameters of the Compared Methods}
\label{app:baselines}
\textbf{Noise-space GRPO (Table~\ref{tab:main5k}):} Our own implementation of noise-space GRPO. The verifier is $\Vact$ (trained for 60k steps), the advantage is computed with the top1 scheme, and the noise is edited by additive perturbation. The initial value of $\log\sigma$ is $-1$, the learning rate is $10^{-4}$, the success filter is applied, and no $\mu$ trust region is used. Adaptation uses $K=10$ (seeds 0--9, softmax-aggregated execution), and evaluation uses seeds 10--59 with both a single candidate (one candidate generated from the adapted mean $\mu$) and $+$\boe (weighted aggregation).
\textbf{Noise-space GRPO and DSRL-style SAC (Table~\ref{tab:budget}):} Adaptation seeds are $0,\ldots,K-1$, and evaluation uses the 50 seeds immediately following, with single-candidate generation. GRPO uses argmax-selection adaptation and the success filter. SAC is a port of the official DSRL implementation (dsrl\_pi0), with noise-space replacement (pure scheme), a $-1/0$ reward ($-1$ per chunk and $0$ at the terminal state upon task success), a TanhNormal distribution clipped to $[-1, 1]$, an ensemble mean over 10 critics, discount factor 0.99, actor / critic / temperature learning rates of $10^{-4}$ / $3\times10^{-4}$ / $3\times10^{-4}$, batch size 16, UTD ratio 1, and updates starting after 300 transitions have accumulated. SAC does not use the verifier.

\subsection{Setup of the Self-Imitation Control}
\label{app:self-imitation}
``Self'' in Table~\ref{tab:main5k} runs the SmolVLA 5k checkpoint with $N=1$ at every replanning boundary.
Because the LoRA output weights are zero-initialized, the data-collection policy is identical to the base policy.
A single input noise $z_0\sim\mathcal{N}(0,I)$ is drawn from the standard normal distribution, and the candidate weight is trivially 1. The verifier is not used at all during the adaptation trials.
The pair of observation $o_b$ and generated chunk $A_b$ at each boundary is stored in a temporary buffer, and after the episode ends, only the pairs from episodes judged successful are added to the distillation buffer $\mathcal D$.
The teacher is the full $H_1=50$ steps in the model's normalized space, of which the first $H_2=10$ steps are executed in the environment.
This corresponds to a control in which the teacher of the proposed method (Distill), the aggregated chunk $\bar A_b$, is replaced by the chunk $A_b$ generated by the base policy alone.
The LoRA is distilled for 200 steps on the successes among the 10 collected episodes, and the 40 tasks are evaluated with seeds 10--59.
For the 3 tasks whose distillation buffer was empty because no success was obtained, the base policy was evaluated without any weight update.
Self is evaluated with single-candidate generation, and ``Self + \boe'' with weighted aggregation of eight candidates using the same $\Vact$ at evaluation time.

\subsection{Compute Environment}
\label{app:compute}
All experimental evaluations were run on NVIDIA RTX PRO 6000 (96 GB) and RTX 6000 Ada (48 GB) GPUs, with 2--4 jobs placed in parallel on a single GPU.
Adaptation for one task (10 episodes of \boe execution + 200 LoRA update steps) takes 5--10 minutes, and 100 evaluation episodes (50 single-candidate + 50 \boe) take 20--40 minutes. One row of the main experiment (Table~\ref{tab:main5k}) over all 40 tasks corresponds to roughly 20--40 GPU-hours.

\subsection{Verifier Architecture and Training}
\label{app:verifier}
The candidate dataset was built by generating eight candidates with the mixed-$\alpha$-scale scheme, per target policy checkpoint, at states from both the official LIBERO demonstrations (all four suites) and the newly collected successful rollouts (104,294 states for the 10k policy dataset).
The supervision is the ranking induced by the environment-space RMSE to the $H_2=10$ recorded action steps of the source trajectory at each state. Training uses an all-pairs Bradley--Terry loss for 60k steps with batch size 16, learning rate $10^{-4}$, weight decay 0.01, 5\% warmup, and gradient clipping at 1.0. The evaluator is a Transformer with hidden size 512 and 6 layers.
The latent evaluator $\Vlat$ uses identical hyperparameters, with only the input replaced by latent variables predicted by the world model.
For each evaluated checkpoint (5k / 10k / 20k), $\Vact$ and $\Vlat$ were trained separately under the same compute budget.
The validation pairwise accuracy at the end of training (60k steps), reported as $\Vact$ / $\Vlat$, was 0.82 / 0.70 for the 5k policy, 0.73 / 0.62 for the 10k policy, and 0.72 / 0.60 for the 20k policy.
This pairwise accuracy measures agreement with the distance-based teacher ranking; the final improvement of the policy is evaluated by the closed-loop success rates reported in the main text.

\subsection{World Model Architecture}
\label{app:wm}
The world model is ``Pi05SplitLatentWorldModel,'' which takes a history of the past 3 latent frames, the proprioceptive state, and a candidate action chunk as input and outputs the predicted latent $\hat{\vz}_{t+H_2}$ after executing the first $H_2$ steps.
It has a two-branch residual structure with an action-conditioned branch and an action-independent branch, feature dimension 512, depth 4 (2 for the independent branch), 8 attention heads, and MLP hidden dimension 2048.
It was trained for 20k steps on the official LIBERO demonstrations and the newly collected rollouts (both successes and failures). The L1 error on validation data is 3.09, a 36\% improvement over the naive baseline that simply copies the previous state (error 4.86), with a cosine similarity of 0.96.

\subsection{Distillation Details}
\label{app:distill-impl}
The LoRA targets follow SmolVLA's standard fine-tuning configuration: the Query/Value projections of the action expert, the state/action input and output projections, and the time MLP layers, for a total of 74 tensors (0.37M parameters). The success buffer collected over $K=10$ episodes averaged 111 boundaries per task in the 5k main experiment and 84 boundaries in the self-imitation control.
Updates were run for 200 steps with batch size 8, the Adam optimizer (learning rate $10^{-4}$), bfloat16 automatic mixed precision (autocast), and gradient-norm clipping at 1.0.

\section{Ablation on the Success Filter and Teacher Selection}\label{app:distill-abl}
Table~\ref{tab:distill-abl} reports ablations that vary the type of teacher chunk, the presence of the success filter, and the verifier input.
Results with single-candidate generation are in Table~\ref{tab:distill-abl}(a) and those with candidate aggregation (eight candidates) in Table~\ref{tab:distill-abl}(b).

\begin{table}[t]
\centering
\small
\setlength{\tabcolsep}{3pt}
\caption{Comparison of distillation teacher, success filter, and verifier (SmolVLA 5k, 10 adaptation episodes, evaluation seeds 10--59, 500 episodes per suite).
Success rates (\%) with (a) single-candidate generation and (b) weighted aggregation of eight candidates.
``Mean'' uses the weighted-average chunk and ``Max'' the highest-scoring candidate as the teacher.
The main-method setting is ``Mean teacher, filter on, $\Vact$.'' The $\Vlat$ rows also use $\Vlat$ for candidate evaluation.}
\label{tab:distill-abl}
\textbf{(a) Single-candidate generation}\par\smallskip
\begin{tabular*}{\linewidth}{@{\extracolsep{\fill}}lccrrrrr@{}}
\toprule
Teacher & Filter & Verifier & Spatial & Object & Goal & Long & Avg. \\
\midrule
Mean & On & $\Vact$ & 81.8 & 88.0 & 79.4 & 33.2 & 70.6 \\
Max & On & $\Vact$ & 78.4 & 86.8 & 78.2 & 27.8 & 67.8 \\
Mean & Off & $\Vact$ & 73.8 & 85.2 & 71.4 & 27.0 & 64.4 \\
Max & Off & $\Vact$ & --- & --- & --- & 32.6 & --- \\
Mean & Off & $\Vlat$ & 62.2 & 84.4 & 67.2 & 25.8 & 59.9 \\
\bottomrule
\end{tabular*}
\par\medskip
\textbf{(b) Candidate aggregation (eight candidates)}\par\smallskip
\begin{tabular*}{\linewidth}{@{\extracolsep{\fill}}lccrrrrr@{}}
\toprule
Teacher & Filter & Verifier & Spatial & Object & Goal & Long & Avg. \\
\midrule
Mean & On & $\Vact$ & 88.2 & 95.4 & 87.8 & 57.8 & 82.3 \\
Max & On & $\Vact$ & 89.4 & 95.4 & 91.2 & 54.0 & 82.5 \\
Mean & Off & $\Vact$ & 83.0 & 94.6 & 82.4 & 55.0 & 78.8 \\
Max & Off & $\Vact$ & --- & --- & --- & 56.2 & --- \\
Mean & Off & $\Vlat$ & 79.2 & 87.0 & 79.4 & 43.8 & 72.4 \\
\midrule
\multicolumn{3}{l}{No adaptation ($\Vlat$)} & 74.0 & 89.8 & 81.2 & 41.6 & 71.7 \\
\bottomrule
\end{tabular*}
\end{table}

Our standard setting is ``aggregated-mean teacher, success filter on, $\Vact$.''
Note that the distillation results with $\Vlat$ were measured without the success filter and should therefore be compared against the corresponding filter-off rows for $\Vact$.

\subsection{Distilling Only the Executed Segment ($H_2$-Only Teacher)}
\label{app:h2only}
The main recipe (\S\ref{sec:tta}) imitates the full $H_1=50$-step aggregated chunk even though the verifier scores, and the environment executes, only its first $H_2=10$ steps.
To test whether the unexecuted $H_1-H_2$ steps are necessary as a teacher, we distilled from the same success buffers as the main experiment (SmolVLA 5k, $K=10$, success filter on, $\Vact$, 200 LoRA steps) while masking the flow-matching loss in Eq.~\eqref{eq:fm} to the first $H_2$ steps of each chunk; the remaining 40 steps contribute no gradient. Everything else, including the evaluation seeds 10--59, is identical.

\begin{table}[h]
\caption{Full-chunk teacher ($H_1=50$, main recipe) vs.\ $H_2$-only teacher (loss masked to the executed 10 steps). Success rates (\%) on LIBERO (clean), SmolVLA 5k, seeds 10--59 ($n{=}500$ per suite). The full-chunk rows are those of Table~\ref{tab:main5k}.}
\label{tab:h2only}
\begin{center}
\small
\begin{tabular}{llccccc}
\toprule
Evaluation & Teacher & Spatial & Object & Goal & Long & Avg. \\
\midrule
\multirow{2}{*}{Single candidate}
 & Full chunk ($H_1$) & \textbf{81.8} & 88.0 & 79.4 & 33.2 & 70.6 \\
 & Executed steps ($H_2$) & 79.0 & \textbf{92.0} & \textbf{84.2} & \textbf{41.0} & \textbf{74.1} \\
\midrule
\multirow{2}{*}{$+$\boe}
 & Full chunk ($H_1$) & 88.2 & 95.4 & 87.8 & 57.8 & 82.3 \\
 & Executed steps ($H_2$) & \textbf{90.4} & \textbf{99.0} & \textbf{92.6} & \textbf{64.2} & \textbf{86.6} \\
\bottomrule
\end{tabular}
\end{center}
\end{table}

As Table~\ref{tab:h2only} shows, restricting the teacher to the executed segment does not hurt; on average it helps.
With single-candidate generation, the $H_2$-only variant reaches 74.1\% against 70.6\% for the full-chunk teacher, with gains on Object ($+4.0$), Goal ($+4.8$), and Long ($+7.8$) and a loss on Spatial ($-2.8$); with candidate aggregation it is higher on every suite (86.6\% vs.\ 82.3\% on average, $+6.4$ on Long).
The unexecuted tail of the aggregated chunk is therefore not needed for retention or re-amplification, and supervising only the steps that the verifier actually scored and the environment actually executed is at least as effective.
A plausible reading is that the tail, which was never validated by execution and is re-planned away at the next boundary, adds noise to the target, whereas the full chunk mainly preserves the dimensions of SmolVLA's pretraining objective.
We keep the full-chunk teacher in the main experiments for consistency with the rest of the paper; these numbers come from a single training seed, so the suite-level differences should be read as indicative, but the $H_2$-only variant is an attractive default and removes the need to imitate unexecuted actions.

\section{Supplementary Experiments on the Number of Candidates and the Adaptation Budget}
\label{app:sweeps}
Table~\ref{tab:nsweep} reports results for varying the number of candidates $N$, and Table~\ref{tab:budget} for varying the number of adaptation episodes $K$.
They correspond to the plots in Figures~\ref{fig:nsweep} and~\ref{fig:budget} in the main text.
Both were run with the variant that does not use the success filter; in the $N$ sweep, the same distilled LoRA is evaluated while varying only the number of candidates at evaluation time.

\begin{table}[t]
\caption{Numbers underlying Figure~\ref{fig:nsweep}: success rates (\%) of \bon\ / distilled policy + \bon, SmolVLA 5k, $n{=}500$ per cell.}
\label{tab:nsweep}
\begin{center}
\begin{tabular}{lccccc}
\toprule
$N$ & Spatial & Object & Goal & Long & Avg. \\
\midrule
1 & 57.6 / \textbf{73.8} & 80.4 / \textbf{85.2} & 61.0 / \textbf{71.4} & 17.4 / \textbf{27.0} & 54.1 / \textbf{64.4} \\
2 & 68.0 / \textbf{76.0} & 82.6 / \textbf{89.0} & 70.6 / \textbf{76.4} & 26.0 / \textbf{37.6} & 61.8 / \textbf{69.8} \\
4 & 74.6 / \textbf{79.0} & 89.8 / \textbf{93.4} & \textbf{81.8} / 81.2 & 36.6 / \textbf{43.6} & 70.7 / \textbf{74.3} \\
8 & 80.6 / \textbf{83.0} & 94.4 / \textbf{94.6} & \textbf{86.8} / 82.4 & 44.0 / \textbf{55.0} & 76.4 / \textbf{78.8} \\
16 & 84.0 / \textbf{86.0} & 92.8 / \textbf{96.4} & \textbf{90.6} / 86.2 & 56.2 / \textbf{62.0} & 80.9 / \textbf{82.7} \\
\bottomrule
\end{tabular}
\end{center}
\end{table}

\begin{table}[t]
\caption{Success rate (\%) on LIBERO-Long as a function of the adaptation-episode budget $K$ ($n{=}500$). Parameters are frozen after $K$ adaptation episodes and then evaluated.
DSRL-style is our port of the official implementation (dsrl\_pi0), with a $-1/0$ reward (time penalty), a TanhNormal distribution restricted to $[-1, 1]$, an ensemble of 10 critics, and prior-replacement composition. The GRPO rows are the comparison method (verifier-ranking reward, top1 advantage, success filter, argmax adaptation).
The distilled policy is the variant without the success filter (Appendix~\ref{app:distill-abl}), with adaptation seeds $0,\ldots,K-1$ and the 50 seeds immediately following as evaluation seeds.
Base performance is 17.4 at 5k and 68.6 at 10k (seeds 10--59); \boe is 44.0 at 5k and 80.0 at 10k.}
\label{tab:budget}
\begin{center}
\begin{tabular}{llccccc}
\toprule
Policy (Base) & Method & $K{=}2$ & $K{=}5$ & $K{=}10$ & $K{=}20$ & $K{=}50$ \\
\midrule
\multirow{4}{*}{5k (17.4)}
 & Distill           & \textbf{29.6} & \textbf{29.2} & \textbf{27.0} & \textbf{29.2} & \textbf{34.6} \\
 & Distill + \boe\        & 52.6 & 52.8 & 55.0 & 53.6 & 54.6 \\
 & N-GRPO       & --- & --- & 25.6 & --- & 29.6 \\
 & DSRL          & --- & --- & 19.4 & --- & 18.8 \\
\midrule
\multirow{4}{*}{10k (68.6)}
 & Distill           & \textbf{75.0} & 70.4 & \textbf{73.8} & \textbf{75.0} & \textbf{75.4} \\
 & Distill + \boe\        & 75.2 & 76.8 & 79.0 & 80.4 & 85.4 \\
 & N-GRPO       & --- & --- & 71.2 & --- & 71.4 \\
 & DSRL          & --- & --- & 70.4 & --- & 70.8 \\
\bottomrule
\end{tabular}
\end{center}
\end{table}

Because the $K$ sweep uses adaptation seeds $0,\ldots,K-1$ and evaluation seeds $K,\ldots,K+49$, the Base numbers evaluated on seeds 10--59 cannot be strictly compared as a common baseline for conditions other than $K=10$.
In addition, although DSRL-style SAC uses the same number of episodes, it does not use the recorded-action ranking supervision used to train the verifier, so the total amount of available supervision is not perfectly matched. Bold indicates the highest success rate in each column.

\section{Details of Retention at the 10k Checkpoint}
\label{app:tab1-10k}
Table~\ref{tab:main10k} summarizes the success rates of the base policy, candidate aggregation, and the distilled policy at the 10k checkpoint.

\begin{table}[h]
\caption{Success rates (\%) on LIBERO (clean) with SmolVLA 10k. Measured under the same protocol as Table~\ref{tab:main5k} (evaluation seeds 10--59, $n{=}500$ per suite).}
\label{tab:main10k}
\begin{center}
\begin{tabular}{llcccc c}
\toprule
Verifier & Method & Spatial & Object & Goal & Long & Avg. \\
\midrule
--- & Base & 85.0 & 95.0 & 84.0 & 68.6 & 83.2 \\
\midrule
\multirow{3}{*}{$\Vact$}
 & \boe             & 90.2 & \textbf{99.6} & \textbf{91.8} & 80.0 & 90.4 \\
 & Distill   & 89.4 & 96.0 & 85.4 & 73.8 & 86.2 \\
 & Distill + \boe  & 91.8 & 98.4 & 89.8 & 79.0 & 89.8 \\
\midrule
\multirow{3}{*}{$\Vlat$}
 & \boe             & \textbf{92.4} & 99.0 & 91.4 & \textbf{82.6} & \textbf{91.4} \\
 & Distill   & 91.0 & 97.0 & 80.8 & 75.6 & 86.1 \\
 & Distill + \boe  & 90.2 & 99.4 & 86.8 & 80.8 & 89.3 \\
\bottomrule
\end{tabular}
\end{center}
\end{table}

As Table~\ref{tab:main10k} shows, the single-candidate success rate of the distilled policy rises from 83.2\% to 86.2\%, so the improvement is reliably retained in inference without additional verifier evaluation.
On the other hand, stacking candidate aggregation on the distilled policy does not exceed the original candidate aggregation (\boe), confirming that retention of improvement and re-amplification of subsequent TTC behave differently.

\section{Verifier Input Representation: Direct Action Evaluation vs.\ World-Model Latents}
\label{app:vlat}
$\Vlat$ is a verifier whose input is the latent state $\hat{\vz}_{t+H_2}$, $H_2$ steps ahead, predicted from the candidate action by the trained world model (Appendix~\ref{app:wm}) in the same $2 \times 64 \times 960$ token space as the observation.
It shares the same encoder, the same candidate sets, the same supervision, the same BT loss, and the same number of training steps (60k) as $\Vact$, differing only in the form of the input representation and in whether the world model is interposed. Table~\ref{tab:selector} compares the two.

\begin{table}[h]
\caption{Comparison of verifier inputs (success rate \%; \boe is softmax aggregation with $\lambda{=}0.3$; each verifier trained for 60k steps).
Clean uses evaluation seeds 10--59 ($n{=}500$ per suite); plus uses 105 tasks $\times$ 10 episodes (seeds 10--19).
$\Delta$ is the per-task paired difference $\Vlat-\Vact$ (mean$\pm$SEM). The comparison as a distillation teacher is based on the variant without the success filter ($K{=}10$, single-candidate generation).}
\label{tab:selector}
\begin{center}
\small
\setlength{\tabcolsep}{3pt}
\begin{tabular*}{\linewidth}{@{\extracolsep{\fill}}lrrrrr@{}}
\toprule
Method & Spatial & Object & Goal & Long & Avg. \\
\midrule
\multicolumn{6}{l}{\textit{clean 5k}} \\
Base                         & 57.6 & 80.4 & 61.0 & 17.4 & 54.1 \\
\boe\ $\Vact$                & \textbf{80.6} & \textbf{94.4} & \textbf{86.8} & \textbf{44.0} & \textbf{76.4} \\
\boe\ $\Vlat$                & 74.0 & 89.8 & 81.2 & 41.6 & 71.7 \\
$\Delta$     & $-6.6 \pm 3.7$ & $-4.6 \pm 1.7$ & $-5.6 \pm 2.5$ & $-2.4 \pm 2.2$ & $-4.8 \pm 1.3$ \\
\midrule
\multicolumn{6}{l}{\textit{clean 5k, as distillation teacher}} \\
Distill ($\Vact$)   & 73.8 & 85.2 & 71.4 & 27.0 & 64.4 \\
Distill ($\Vlat$)   & 62.2 & 84.4 & 67.2 & 25.8 & 59.9 \\
$\Delta$     & $-11.6 \pm 4.3$ & $-0.8 \pm 4.1$ & $-4.2 \pm 3.5$ & $-1.2 \pm 1.2$ & $-4.5 \pm 1.8$ \\
\midrule
\multicolumn{6}{l}{\textit{clean 10k}} \\
Base                         & 85.0 & 95.0 & 84.0 & 68.6 & 83.2 \\
\boe\ $\Vact$                & 90.2 & \textbf{99.6} & \textbf{91.8} & 80.0 & 90.4 \\
\boe\ $\Vlat$                & \textbf{92.4} & 99.0 & 91.4 & \textbf{82.6} & \textbf{91.4} \\
$\Delta$     & $+2.2 \pm 1.5$ & $-0.6 \pm 0.6$ & $-0.4 \pm 2.7$ & $+2.6 \pm 1.9$ & $+0.9 \pm 0.9$ \\
\midrule
\multicolumn{6}{l}{\textit{clean 10k, as distillation teacher}} \\
Distill ($\Vact$)   & 89.4 & 96.0 & 85.4 & 73.8 & 86.2 \\
Distill ($\Vlat$)   & 91.0 & 97.0 & 80.8 & 75.6 & 86.1 \\
$\Delta$     & $+1.6 \pm 2.7$ & $+1.0 \pm 1.2$ & $-4.6 \pm 2.3$ & $+1.8 \pm 3.2$ & $-0.1 \pm 1.3$ \\
\midrule
\multicolumn{6}{l}{\textit{plus 10k}} \\
Base                         & 35.5 & 18.8 & 13.0 & 16.2 & 20.9 \\
\boe\ $\Vact$                & 41.5 & 24.9 & \textbf{14.3} & \textbf{21.7} & 25.6 \\
\boe\ $\Vlat$                & \textbf{43.6} & \textbf{32.5} & 13.0 & 20.6 & \textbf{27.4} \\
$\Delta$     & $+2.1 \pm 1.6$ & $+7.6 \pm 1.7$ & $-1.2 \pm 1.1$ & $-1.1 \pm 1.8$ & $+1.8 \pm 0.8$ \\
\bottomrule
\end{tabular*}
\end{center}
\end{table}

\begin{figure}[h]
\begin{center}
\includegraphics[width=0.5\linewidth]{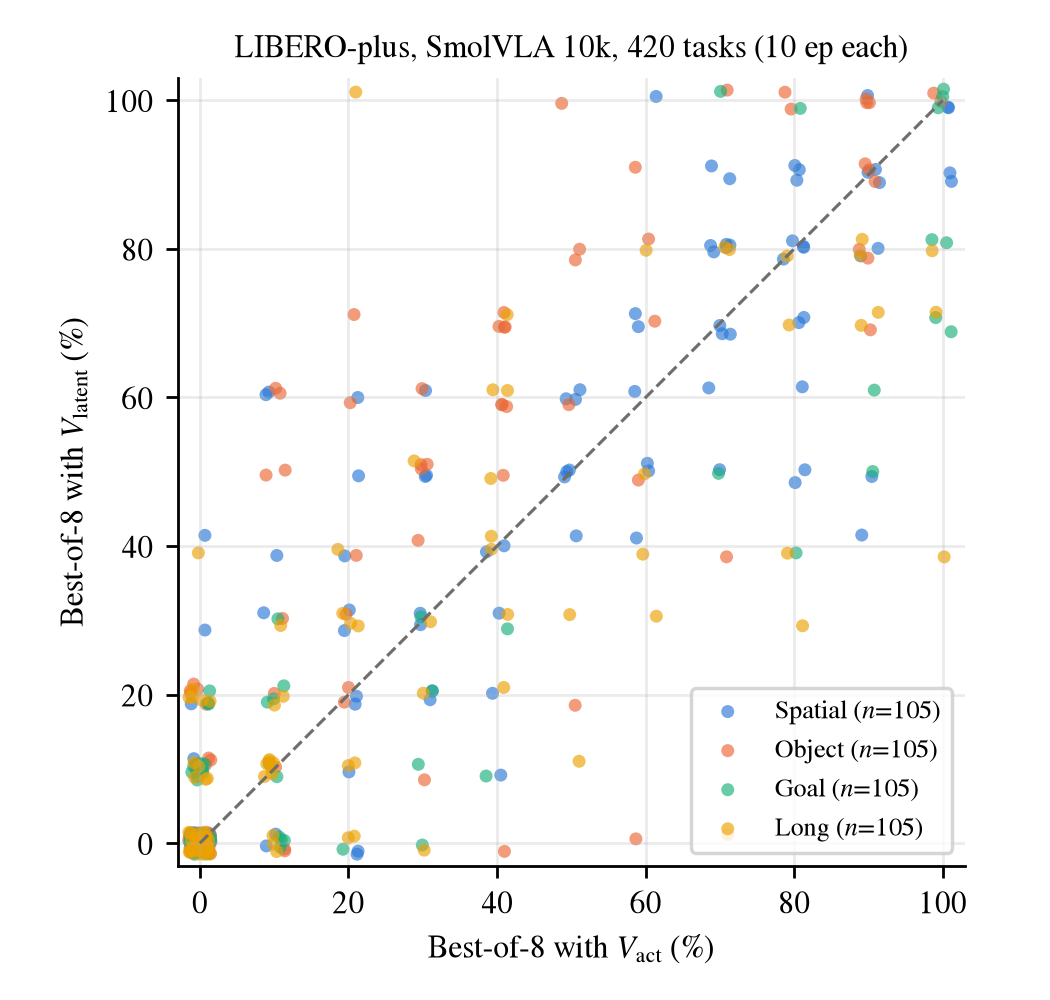}
\end{center}
\caption{Scatter plot comparing \boe success rates on all 420 LIBERO-plus tasks (SmolVLA 10k): $\Vact$ (horizontal) vs.\ $\Vlat$ (vertical), 10 episodes per point
(jitter of $\pm1.5$ points added to reduce overlap). Points above the diagonal favor $\Vlat$. The upward shift is especially pronounced on the Object suite.}
\label{fig:vlat}
\end{figure}

The improved/tied/degraded task counts corresponding to the differences $\Delta$ in Table~\ref{tab:selector} are 7/9/24 on clean 5k, 16/13/11 on clean 10k, and 127/210/83 on LIBERO-plus for aggregated execution, and 14/4/22 on clean 5k and 15/8/17 on clean 10k for the distilled-policy evaluation.
As Table~\ref{tab:selector} and Figure~\ref{fig:vlat} show, $\Vlat$ yields a clear improvement on LIBERO-plus, particularly on the Object suite.
However, because introducing the world model changes both the abstraction level of the input representation and the model structure, we have not yet identified the detailed mechanism behind this advantage.

\section{Encoder and Feature-Space Ablation}
\label{app:encoder}
The feature-space comparison in this section uses a SmolVLA 4k checkpoint trained on successful rollouts from the Object and Long suites.
The success rate of this base policy is 81\% on Object and 36\% on Long.
Table~\ref{tab:encoder-protocol} summarizes the differences between this setup and the main experiments, which use the 5k/10k/20k checkpoints trained on all four suites.

\begin{table}[t]
\centering
\small
\caption{Differences in setup between the feature-space comparison (Figure~\ref{fig:encoder}) and the main experiments (clean LIBERO). ``a1.2'' in the candidate-dataset names denotes the noise multiplier.}
\label{tab:encoder-protocol}
\begin{tabular}{p{0.16\linewidth}p{0.37\linewidth}p{0.37\linewidth}}
\toprule
Item & Feature-space comparison (Figure~\ref{fig:encoder}) & Main experiments \\
\midrule
Policy & BC-trained on the two suites Object and Long. 4k checkpoint (\path{run_bc_success/checkpoints/004000}) & BC-trained on successful rollouts from all four suites. 5k / 10k / 20k checkpoints of \path{run_bc_4suite} \\
Candidate data & \path{cand_4k_a1.2}. Object and Long candidates generated from the 4k policy & \path{cand_4suite_<ckpt>_a1.2}. Four suites, generated per policy checkpoint \\
Verifier training & 20k steps & 60k steps \\
World model & \path{run1} (SmolVLA features) and \path{run_dinov2} (DINOv2 features) & \path{run_4suite} (SmolVLA features, all four suites) \\
Candidate noise at evaluation & $\alpha=2.0$ & $\alpha=1.2$ \\
Execution rule & Select the highest-scoring of 8 candidates (argmax) & Softmax-weighted aggregation of 8 candidates ($\lambda=0.3$) \\
Evaluation scale & Object and Long, 25 episodes per task (250 per suite). Seed set differs from the main experiments & All four suites, 50 episodes per task (500 per suite), seeds 10--59 \\
Feature encoder & 20k checkpoint of \path{run_bc_success} & 20k checkpoint of \path{run_bc_4suite} \\
\bottomrule
\end{tabular}
\end{table}

\begin{figure}[h]
\begin{center}
\includegraphics[width=0.8\linewidth]{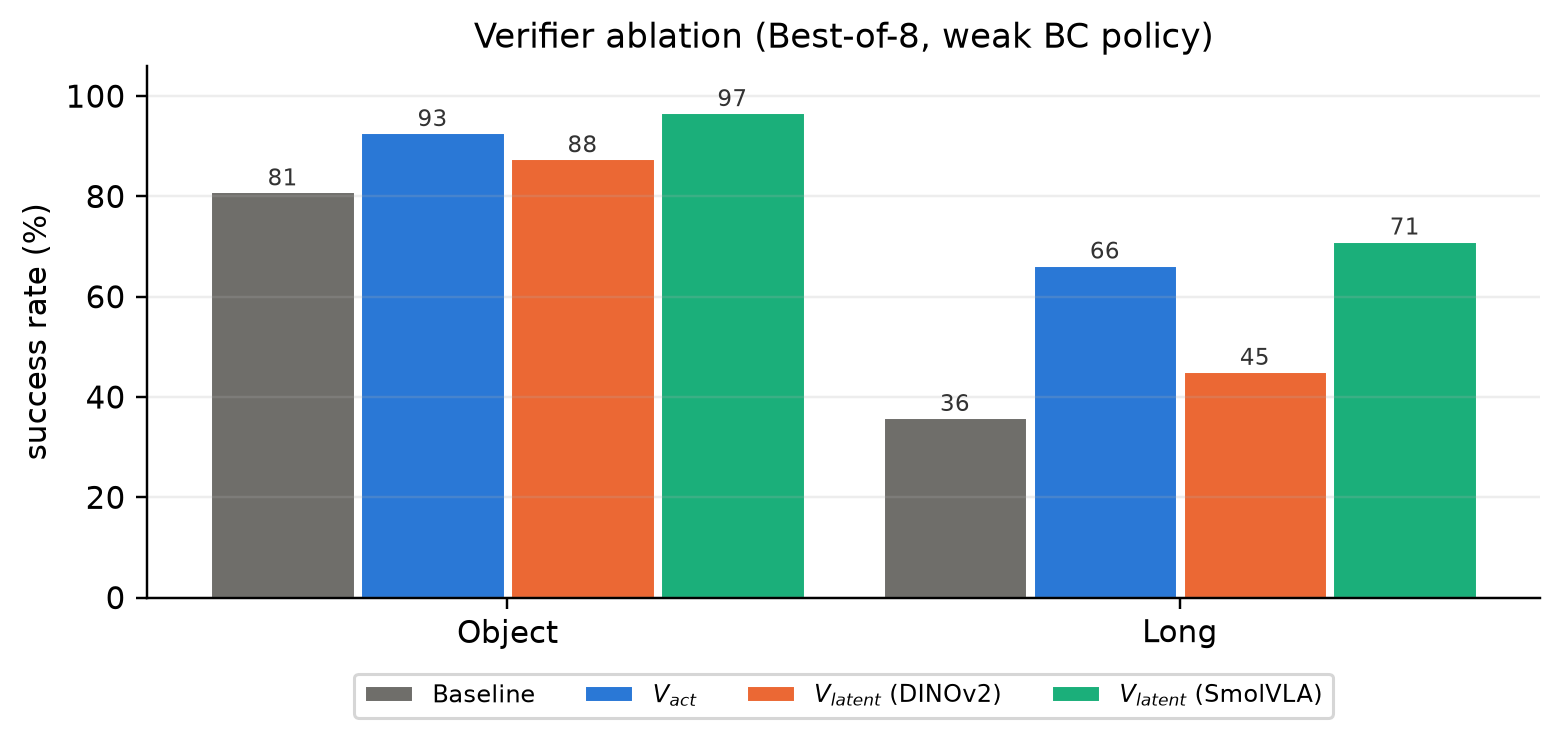}
\end{center}
\caption{Comparison of verifier input representations with a SmolVLA 4k checkpoint trained on the Object and Long suites.
Eight candidates, noise scale $\alpha=2.0$, argmax selection, 25 evaluation episodes per task (250 per suite).
Verifiers are trained for 20k steps. The base policy, $\Vact$ with direct action input, and $\Vlat$ predicting DINOv2 features or SmolVLA features through the world model are compared.
See Table~\ref{tab:encoder-protocol} for details of the setup.}
\label{fig:encoder}
\end{figure}

The verifier's candidate dataset was generated with $\alpha=1.2$, whereas the closed-loop evaluation in Figure~\ref{fig:encoder} used $\alpha=2.0$.
The feature-space comparison was also evaluated separately under the $\alpha=1.2$ condition.

In Figure~\ref{fig:encoder}, the success rates on Object / Long are 81\% / 36\% for the base policy, 93\% / 66\% for $\Vact$,
88\% / 45\% for $\Vlat$ with DINOv2 features, and 97\% / 71\% for $\Vlat$ with SmolVLA features.
Under this comparison condition, candidate evaluation through a world model that predicts SmolVLA's own features achieved the highest success rates.
Although DINOv2 $\ge$ SmolVLA held in every individual diagnostic evaluation of frame-level perturbation invariance, state-decoding accuracy, and the world model's action sensitivity, the ranking of these offline feature diagnostics did not match the ranking of final closed-loop success rates.

\section{Softmax-Aggregated Execution vs.\ Argmax Selection}
\label{app:softmax}
On the Long suite with the 5k checkpoint, evaluated with $\Vact$ over 50 episodes per task (seeds 0--49), argmax selection (\selEight) achieved a success rate of 34.4\%, whereas softmax aggregation ($\lambda=0.3$, \boe) achieved 44.2\% and softmax aggregation with $\Vlat$ achieved 43.2\%.
On the Long suite with the 10k checkpoint, argmax selection gave 77.0\%, softmax aggregation 78.2\%, and softmax aggregation with $\Vlat$ 82.2\%.
When GRPO adapted under argmax selection was evaluated with softmax aggregation, its success rate on 5k Long was 51.8\% (53.4\% under argmax evaluation).

\section{Effect of the Candidate Noise Scale $\alpha$}
\label{app:alpha}
Table~\ref{tab:alpha} shows success rates on the Object and Long suites as the noise multiplier $\alpha$ used for candidate generation is varied.
For every checkpoint, $\alpha=1.2$ achieved the most consistently high success rate within the explored range $\{1.2, 1.5, 2.0, 3.0\}$.

\begin{table}[h]
\caption{Success rate as a function of the candidate noise scale $\alpha$ ($N{=}8$, $\Vlat$, $n{=}100$ per point, Long / Object suites). Bold marks the maximum in each row including the base policy (ties included). The explored range is $\{1.2, 1.5, 2.0, 3.0\}$.}
\label{tab:alpha}
\begin{center}
\begin{tabular}{llccccc}
\toprule
Checkpoint & Suite & Base & $\alpha{=}1.2$ & $1.5$ & $2.0$ & $3.0$ \\
\midrule
5k & Object & \textbf{82} & \textbf{82} & 59 & 39 & 44 \\
5k & Long   & 17 & \textbf{42} & 21 & 7 & 6 \\
10k & Object & 96 & \textbf{98} & 93 & 64 & 71 \\
10k & Long   & 67 & \textbf{74} & 60 & 27 & 21 \\
20k & Object & \textbf{99} & 98 & 91 & 84 & 75 \\
20k & Long   & \textbf{90} & 79 & 68 & 33 & 30 \\
\bottomrule
\end{tabular}
\end{center}
\end{table}

\section{Detailed Formulation of Noise-Space GRPO}
\label{app:grpo}

\subsection{Formulation}
N-GRPO edits the initial noise fed to the frozen VLA with a state-dependent perturbation vector $\eta$.
The noise-editing policy takes as input a 1928-dimensional vector formed by concatenating patch-averaged features from the frozen observation encoder for the two views ($2\times960$ dimensions) and the normalized proprioceptive state (8 dimensions).
A 2-layer MLP with hidden width 64 outputs a 32-dimensional mean vector $\mu$ (zero-initialized), which together with a state-independent variance parameter $\log\sigma$ (initial value $-1$) defines a diagonal multivariate Gaussian. A perturbation $\eta$ sampled from this distribution is broadcast along the time axis of the action chunk and added.
Among the generated candidates, candidate 0 is an ordinary sample without perturbation, and candidates $i=1,\ldots,7$ are formed by adding the same $\eta_i$ to independently generated noise.
The advantage uses the top1 scheme ($+1$ for the candidate with the highest verifier score and $-1/(N-1)$ for the others), and updates use learning rate $10^{-4}$, PPO clip ratio 0.2, replay buffer size 256, and 3 epochs.
As update constraints, we introduce a ``success filter,'' which defers updates until the end of the episode and applies them only upon success, and a ``$\mu$ trust region,'' which restricts the parameter norm to $\|\mu\| \le 1$.

The above is the basic configuration of the comparison implementation; the main experiment in Table~\ref{tab:main5k} does not use the $\mu$ trust region, and the adaptation-budget experiment (Table~\ref{tab:budget}) uses argmax-selection adaptation.

\section{Reproducibility of the Correction Direction under Re-Sampling of the Candidate Set}
\label{app:amort}
We examined how consistent the verifier's correction direction is when the candidate set is independently re-drawn at the same state.
We loaded 32 fixed LIBERO-plus states (8 per suite, extracted from diag80 rollouts) at each of the 5k / 10k / 20k checkpoints and independently re-sampled the candidate set ($N=8$, $\alpha=1.2$) $M=32$ times at each state.
Let $A^{(i)}_m$ be the candidates obtained in the $m$-th draw, $\langle A \rangle_m = \frac{1}{N}\sum_i A^{(i)}_m$ their simple mean, and $w_{m,i}$ the softmax weights;
we define the TTC correction vector as $\Delta_m = \sum_i w_{m,i}(A^{(i)}_m - \langle A \rangle_m)$ (350 dimensions = $H_1=50$ steps $\times$ 7 environment-action dimensions).
We computed the following metrics:
(i) the normalized resultant length $R = \|\frac{1}{M}\sum_m \Delta_m / \|\Delta_m\|\|$ (a measure of directional agreement whose expected value for completely uncorrelated directions is about $1/\sqrt{M} \approx 0.18$),
(ii) the split-half cosine similarity (the cosine similarity between the mean correction vectors of a random two-way split of the $M$ sets, averaged over 10 trials),
(iii) SDR = $\|\frac{1}{M}\sum_m \Delta_m\| / \frac{1}{M}\sum_m \|\Delta_m\|$ (the ratio of the norm of the mean correction vector to the mean of the norms),
and (iv) the effective number of candidates $N_{\mathrm{eff}} = 1/\sum_i w_{m,i}^2$ (the number of candidates effectively averaged by the softmax).

The measurements (5k / 10k / 20k) were: resultant length $R$ of 0.229 / 0.228 / 0.209, split-half cosine similarity of 0.22 / 0.22 / 0.13, SDR of 0.053 / 0.053 / 0.043, and an effective number of candidates $N_{\mathrm{eff}} \approx 2.1$ in all conditions (Table~\ref{tab:amort}).
Bootstrap 95\% confidence intervals for the paired differences between checkpoints at the same states (5k$-$20k and 10k$-$20k) excluded zero for every metric, indicating significant differences.
In contrast, the split-half Frobenius cosine similarity of the second-moment structure (the weighted covariance $D_m = \sum_i (w_{m,i} - 1/N)\, e_{m,i} e_{m,i}^\top$ with $e_{m,i} = A^{(i)}_m - \langle A \rangle_m$) was $+0.02$ (the null reference with shuffled weights is 0) and showed no difference between checkpoints.
These results reveal that at the most thoroughly trained 20k checkpoint, the variability of the correction direction across re-draws of the candidates at inference time increases markedly compared with 5k and 10k.
The paired differences of 5k minus 20k (bootstrap 95\% CI) were $+0.020$ [0.006, 0.034] for $R$, $+0.090$ [0.037, 0.143] for the split-half cosine similarity, and $+0.010$ [0.003, 0.019] for SDR.
Likewise, the differences of 10k minus 20k were $+0.020$ [0.011, 0.029], $+0.087$ [0.047, 0.127], and $+0.010$ [0.005, 0.015], respectively (the null reference for $R$ is 0.184).

\begin{table}[t]
\centering
\small
\caption{Within-state reproducibility of the TTC correction vector $\Delta_m$ (32 states, 32 candidate-set draws per state).
(a) Directional reproducibility metrics; (b) correction norm, effective number of candidates, and score standard deviation.
$R$ is the resultant length, Split-cos the split-half cosine similarity, and SDR the directional stability ratio.}
\label{tab:amort}
\textbf{(a) Directional reproducibility of the correction vector}\par\smallskip
\begin{tabular*}{\linewidth}{@{\extracolsep{\fill}}lccc@{}}
\toprule
Checkpoint & $R$ & Split-cos & SDR \\
\midrule
5k & $0.229\pm0.007$ & $0.219\pm0.025$ & $0.053\pm0.004$ \\
10k & $0.228\pm0.007$ & $0.216\pm0.027$ & $0.053\pm0.004$ \\
20k & $0.209\pm0.005$ & $0.129\pm0.023$ & $0.043\pm0.002$ \\
\bottomrule
\end{tabular*}
\par\medskip
\textbf{(b) Correction norm and candidate statistics}\par\smallskip
\begin{tabular*}{\linewidth}{@{\extracolsep{\fill}}lccc@{}}
\toprule
Checkpoint & $|\Delta|/\mathrm{spread}$ & $N_{\mathrm{eff}}$ & Score std.\ dev. \\
\midrule
5k & 0.649 & 2.13 & 2.65 \\
10k & 0.658 & 2.11 & 1.60 \\
20k & 0.647 & 2.14 & 1.36 \\
\bottomrule
\end{tabular*}
\end{table}

\section{Behavior of Distillation and Candidate Aggregation at the 20k Checkpoint}
\label{app:strength}

\begin{table}[t]
\centering
\small
\setlength{\tabcolsep}{4pt}
\caption{Success rates (\%) on LIBERO-plus with SmolVLA 20k and per-task paired differences from Base (points, mean$\pm$SEM).
Evaluation seeds 10--19. (a) Aggregated over the 837 Level-5 tasks on which Base has a positive success rate; (b) aggregated over a fixed set of 80 tasks (32 tasks for PRO only).
VWFM is advantage-weighted flow matching over all candidates, Online is online LoRA updating, and PRO is preference optimization (Appendix~\ref{app:20k}).}
\label{tab:plus20k}
\textbf{(a) All 837 tasks}\par\smallskip
\begin{tabular*}{\linewidth}{@{\extracolsep{\fill}}lccc@{}}
\toprule
Method & Success & $\Delta$ Base & $\Delta$ \boe \\
\midrule
Base & $37.3\pm1.1$ & --- & $-6.6\pm0.7$ \\
\boe & $43.9\pm1.2$ & $+6.6\pm0.7$ & --- \\
N-GRPO + \boe & $43.8\pm1.2$ & $+6.5\pm0.7$ & $-0.1\pm0.6$ \\
\bottomrule
\end{tabular*}
\par\medskip
\textbf{(b) Fixed 80-task subset}\par\smallskip
\begin{tabular*}{\linewidth}{@{\extracolsep{\fill}}lccc@{}}
\toprule
Method & Success & $\Delta$ Base & $\Delta$ \boe \\
\midrule
Base & $38.8\pm3.7$ & --- & $-10.1\pm2.2$ \\
\boe & $48.9\pm3.7$ & $+10.1\pm2.2$ & --- \\
N-GRPO + \boe & $46.1\pm3.7$ & --- & $-2.8\pm1.9$ \\
\midrule
Distill & $45.6\pm1.8$ & $+6.9\pm2.1$ & $-3.3\pm2.1$ \\
Distill + \boe & $45.2\pm1.8$ & $+6.4\pm2.2$ & $-3.6\pm2.1$ \\
VWFM & 42.9 & $+4.1\pm2.3$ & --- \\
VWFM + \boe & 49.1 & --- & $+0.2\pm2.1$ \\
Online & 42.0 & $+3.2\pm2.1$ & --- \\
Online + \boe & 49.0 & --- & $+0.1\pm1.8$ \\
PRO & --- & $+3.4\pm2.8$ & --- \\
PRO + \boe & --- & --- & $-0.3\pm2.5$ \\
\bottomrule
\end{tabular*}
\end{table}

The effect size of test-time compute (TTC) varies greatly with the training stage (checkpoint) of the base policy (see Tables~\ref{tab:main5k}, \ref{tab:main10k}, and~\ref{tab:plus20k}: on 5k Long, \boe improves performance by $+26.6$ points, whereas on 20k Long it degrades performance by $-11$ points).
Yet when this 20k checkpoint is evaluated on LIBERO-plus, with its unseen perturbations, \boe again shows a clear improvement of $+6.6$ points ($n=837$).
Nevertheless, the additional gain from stacking candidate aggregation on top of an adaptation method remained very small
(Table~\ref{tab:plus20k}: for GRPO, softmax distillation, advantage-weighted flow matching, online LoRA, and preference optimization alike, the additional effect of combined aggregation was limited, and for the distillation methods it even decreased).
On the other hand, the single-candidate performance of the distilled policy exceeds the base policy by $+6.9 \pm 2.1$ points (positive/zero/negative: 36/29/15), confirming that about 70\% of the gain obtained through TTC ($+10.1$ points) is reliably transferred to lightweight single-candidate inference that requires no verifier.

To probe the background of this phenomenon, we analyzed in detail the within-state reproducibility of the verifier's correction direction, as described above (Appendix~\ref{app:amort}).
When the candidate set was re-sampled $M=32$ times at the same state, the split-half cosine similarity of the correction vector $\Delta_m$ was 0.22 at 5k and 0.22 at 10k, but dropped to 0.13 at 20k.
This suggests that at highly trained checkpoints, the correction direction varies widely across re-draws of the candidates, which may be one factor that hinders the synergy from candidate aggregation after adaptation.

\section{Additional Experiments on Noise-Space GRPO}\label{app:grpo-eval}
\subsection{Evaluation on Clean LIBERO with Argmax Selection}
Table~\ref{tab:grpo-t1} evaluates with the argmax scheme (\selEight), which selects and executes the highest-scoring candidate at each replanning point. The evaluation seeds for the base policy and the candidate-selection policy are 0--49; GRPO uses adaptation seeds 0--9 and evaluation seeds 10--59.

\begin{table}[h]
\caption{GRPO success rates (\%) on clean LIBERO, 50 episodes per task. Base and \selEight use evaluation seeds 0--49 (argmax selection); the GRPO rows use adaptation seeds 0--9 and evaluation seeds 10--59.}
\label{tab:grpo-t1}
\begin{center}
\begin{tabular}{llcccc}
\toprule
Checkpoint & Method & Spatial & Object & Goal & Long \\
\midrule
\multirow{4}{*}{SmolVLA 5k}
 & Base            & 62.6 & 80.0 & 59.6 & 17.4 \\
 & \selEight  & 73.6 & 88.2 & \textbf{81.8} & 34.4 \\
 & N-GRPO      & 59.8 & 83.8 & 68.2 & 23.2 \\
 & N-GRPO+\selEight       & \textbf{80.2} & \textbf{88.2} & 78.8 & \textbf{53.4} \\
\midrule
\multirow{4}{*}{SmolVLA 10k}
 & Base            & 85.0 & 96.4 & 84.0 & 65.6 \\
 & \selEight  & \textbf{89.6} & \textbf{97.8} & \textbf{86.8} & \textbf{77.0} \\
 & N-GRPO      & 85.0 & 96.2 & 86.0 & 69.4 \\
 & N-GRPO+\selEight       & 84.0 & 96.8 & 85.8 & 73.0 \\
\bottomrule
\end{tabular}
\end{center}
\end{table}

\subsection{Evaluation of the 10k Checkpoint on LIBERO-plus}
Table~\ref{tab:grpo-plus} reports the LIBERO-plus success rates of noise-space GRPO for the 10k checkpoint.
We evaluate single-candidate generation and combined argmax selection, and also compare the type of verifier used for candidate evaluation ($\Vact$ vs.\ $\Vlat$).

\begin{table}[h]
\caption{GRPO success rates (\%) on LIBERO-plus 10k, 105 tasks $\times$ 10 episodes, argmax selection (\selEight). GRPO applies the success filter and the $\mu$ trust region. Parentheses indicate the verifier used for the reward and for candidate selection.}
\label{tab:grpo-plus}
\begin{center}
\begin{tabular}{lccccc}
\toprule
Method & Spatial & Object & Goal & Long & Avg. \\
\midrule
N-GRPO ($\Vact$)              & 35.9 & 21.6 & 13.8 & 18.0 & 22.3 \\
N-GRPO+\selEight\ ($\Vact$)             & 42.1 & 29.6 & \textbf{15.3} & \textbf{21.3} & 27.1 \\
N-GRPO+\selEight\ ($\Vlat$)  & 42.4 & 32.5 & 13.4 & 16.7 & 26.3 \\
\bottomrule
\end{tabular}
\end{center}
\end{table}

\subsection{Analysis of Degradation Causes and Protective Mechanisms}
We found two main pathways for performance degradation during adaptation.
(1) Harmful learning from failed episodes: introducing the success filter and updating only from successful trials completely prevents this degradation (single-candidate 5k Object improves from 75.4\% to 83.8\%, and 10k Long $+$\selEight from 63.4\% to 73.0\%). On 5k Long, where the success rate is only about 20\%, the adaptation gain was preserved even though 80\% of the data were discarded for the update (52.6\% $\to$ 53.4\%).
(2) Excessive divergence of the noise mean $\mu$: this occurs specifically when $\Vlat$ is used as the reward ($\|\mu\|$ diverges to 2.7), and introducing the $\mu$ trust-region constraint reliably suppresses the divergence (success rate 18\% $\to$ 63\%).
$\Vact$ tends to select directions that pull distributional drift back toward the center (restoring coefficient $-0.035 \to -0.053$, growing stronger as the policy degrades), whereas the restoring force of $\Vlat$ is only about 40\% of that and does not act strongly even when the policy has degraded.
We also examined a KL regularization term (coefficient $\beta_{\mathrm{KL}} \in \{0.03, 0.1, 0.3\}$), but it over-constrains the useful adaptation range and substantially reduces the adaptation gain (on 5k Long, the success rate drops from 53\% to 46--49\%).

\subsection{Comparison of Advantage Computation Schemes}
Table~\ref{tab:grpo-advantage} compares four advantage computation schemes for noise-space GRPO: ``top1,'' ``zscore,'' ``rank,'' and ``tophalf.''
The comparison was run on 5k Long in clean LIBERO and on the Object and Long suites of LIBERO-plus.

\begin{table}[h]
\caption{Comparison of advantage computation schemes ($n{=}500$ / $n{=}1050$). Assignment schemes that dilute the weight of the top candidates tend to weaken the correct update direction.
Settings with $\Vlat$ reward and the $\mu$ trust region were also tested.}
\label{tab:grpo-advantage}
\begin{center}
\begin{tabular}{lcccc}
\toprule
Advantage scheme & \shortstack{Clean 5k\\Long} & \shortstack{Plus Object\\$\Vact$} & \shortstack{Plus Object\\$\Vlat$} & \shortstack{Plus Long\\$\Vlat$} \\
\midrule
top1     & \textbf{53.4} & 29.6 & 32.5 & \textbf{16.7} \\
zscore   & 52.6 & \textbf{30.0} & 30.8 & 16.4 \\
rank     & 48.6 & 28.0 & 31.7 & 15.4 \\
tophalf  & 48.4 & 27.4 & \textbf{33.0} & 14.7 \\
\bottomrule
\end{tabular}
\end{center}
\end{table}

\subsection{Reproducibility of the Selected Input Noise and Output Actions}
Over 25 states $\times$ 16 repetitions, the pairwise cosine similarity of the optimal noise vectors $\vz^\ast$ selected by the verifier was 0.007 at 5k and 0.001 at 20k, and the resultant lengths were 0.26 / 0.25 (the expected value for random direction vectors is 0.25). Thus, at both checkpoints the directional agreement of the selected noise itself was at the level of random vectors.
However, in the real-environment action space produced through that noise, the selected actions agreed closely with one another (cosine similarity 0.59 at 5k and 0.37 at 20k).
The immediate improvement in the expected verifier score at boundaries where a policy update occurred was only $+0.028$ at 5k and $+0.011$ at 20k (less than 1\% of the within-set score variance), so the local score improvement itself was very small.
The update-dynamics indicators (coherence 0.72 / 0.72, gradient-clip rate, probability ratio, and the evolution of $\log\sigma$) were nearly identical between the two checkpoints.

\section{Comparison of Adaptation Methods at the 20k Checkpoint on LIBERO-plus}
\label{app:20k}
Paired comparisons were run with evaluation seeds 10--19 on the fixed 80-task set (Table~\ref{tab:plus20k}) or on subsets of 16--32 tasks.
\begin{itemize}
\item \textbf{Distillation-based methods:} Distilling the softmax-aggregated chunk improved single-candidate generation by $+6.0 \pm 2.1$ points, advantage-weighted distillation (one sample drawn) by $+3.9 \pm 2.0$ points, and weighted distillation over all candidates (exact) by $+4.1 \pm 2.3$ points, but in every case the additional gain from combined candidate aggregation ($+$\boe) was at most $+0.2$ points.
\item \textbf{Online LoRA updating:} A scheme that performs a one-pass update over the cumulative buffer each time a successful episode is obtained improved single-candidate generation by $+3.2 \pm 2.1$ points and $+$\boe by $+0.1 \pm 1.8$ points. An attempt to also update the verifier online with the new candidates during adaptation caused a large degradation of $-5.0 \pm 3.0$ points.
\item \textbf{Preference optimization:} A method using a FlowPRO-style objective ($\beta=3.5$, $L_{\mathrm{PRO}}$ only, 32 tasks) showed no effect, with a $+$\boe improvement of $-0.0 \pm 2.8$ points.
\item \textbf{Iterative distillation:} An iterative scheme that regenerates the teacher data in each of 4 rounds yielded only $+1.2 \pm 6.0$ points (16 tasks), inferior to a single batch distillation.
\item \textbf{Many candidates ($N=64$):} Increasing the number of candidates to 64 improved the offline oracle-selection performance by $+26.5\%$, but the actual closed-loop TTC success rate decreased by $-5.5\%$, and combined with TTA it decreased by $-4.7 \pm 4.2$ points. Noise sampling by diagonal Gaussian expansion (RoboMonkey-style, $K=64$) did not improve even the offline oracle, because the temporal structure of the chunk collapses.
\item \textbf{Other RL variants:} We tested Flow-GRPO-style FPO (using the CFM loss ratio), Neighbor GRPO (introducing a locally geometry-normalized distance, $+1.9 \pm 4.6$ points), and a candidate-set hybrid (mixing the base policy with the distillation delta, $-0.6 \pm 2.9$ points), but none produced a significant improvement.
\end{itemize}

\section{Evaluation Protocol Details}
\label{app:protocol}
For paired differences $d_j$ between two methods on the same task, the mean $\bar d$ and its standard error over $T$ tasks are
\begin{equation}
 \operatorname{SEM}(\bar d)=\frac{s_d}{\sqrt T},\qquad
 s_d^2=\frac{1}{T-1}\sum_{j=1}^{T}(d_j-\bar d)^2 .
 \label{eq:sem}
\end{equation}
Alongside the mean difference we report the number of tasks that improved / were unchanged / degraded (positive/zero/negative).
The binomial standard errors shown as error bars in the figures treat each episode as independent; they do not account for within-task correlation between trials or for uncertainty due to random seeds.
The maximum episode length is 280 environment steps for Spatial and Object, 300 for Goal, and 520 for Long, and success is determined by the environment's native criterion. Candidate generation and distillation use torch seed 0; variance across training seeds is not evaluated.

\end{document}

%% file: math_commands.tex
\usepackage{amsmath,amsfonts,bm}

\def\eqref#1{equation~\ref{#1}}

\def\1{\bm{1}}

\def\vz{{\bm{z}}}

\DeclareMathAlphabet{\mathsfit}{\encodingdefault}{\sfdefault}{m}{sl}
\SetMathAlphabet{\mathsfit}{bold}{\encodingdefault}{\sfdefault}{bx}{n}



%% file: iclr2027_conference.bib
@misc{chen_sarm2_2026,
	title = {{SARM2}: {Multi}-{Task} {Stage} {Aware} {Reward} {Modeling} for {Self} {Improving} {Robotic} {Manipulation}},
	shorttitle = {{SARM2}},
	url = {http://arxiv.org/abs/2606.10305},
	doi = {10.48550/arXiv.2606.10305},
	language = {en},
	urldate = {2026-09-25},
	publisher = {arXiv},
	author = {Chen, Qianzhong and Zheng, Hau and Yu, Justin and Huang, Suning and Sun, Jiankai and Goldberg, Ken and Wen, Chuan and Abbeel, Pieter and Shentu, Yide and Wu, Philipp and Schwager, Mac},
	month = jun,
	year = {2026},
	note = {arXiv:2606.10305 [cs.RO]},
}

@misc{choi_scale_2026,
	title = {{SCALE}: {Self}-uncertainty {Conditioned} {Adaptive} {Looking} and {Execution} for {Vision}-{Language}-{Action} {Models}},
	shorttitle = {{SCALE}},
	url = {http://arxiv.org/abs/2602.04208},
	doi = {10.48550/arXiv.2602.04208},
	language = {en},
	urldate = {2026-09-25},
	publisher = {arXiv},
	author = {Choi, Hyeonbeom and Ahn, Daechul and Lee, Youhan and Kang, Taewook and Cho, Seongwon and Choi, Jonghyun},
	month = aug,
	year = {2026},
	note = {arXiv:2602.04208 [cs.RO]},
}

@misc{snell_scaling_2024,
	title = {Scaling {LLM} {Test}-{Time} {Compute} {Optimally} can be {More} {Effective} than {Scaling} {Model} {Parameters}},
	url = {http://arxiv.org/abs/2408.03314},
	doi = {10.48550/arXiv.2408.03314},
	language = {en},
	urldate = {2026-09-25},
	publisher = {arXiv},
	author = {Snell, Charlie and Lee, Jaehoon and Xu, Kelvin and Kumar, Aviral},
	month = aug,
	year = {2024},
	note = {arXiv:2408.03314 [cs.LG]},
}

@misc{kwok_scaling_2026,
	title = {Scaling {Verification} {Can} {Be} {More} {Effective} than {Scaling} {Policy} {Learning} for {Vision}-{Language}-{Action} {Alignment}},
	url = {http://arxiv.org/abs/2602.12281},
	doi = {10.48550/arXiv.2602.12281},
	language = {en},
	urldate = {2026-09-25},
	publisher = {arXiv},
	author = {Kwok, Jacky and Zhang, Xilun and Xu, Mengdi and Liu, Yuejiang and Mirhoseini, Azalia and Finn, Chelsea and Pavone, Marco},
	month = feb,
	year = {2026},
	note = {arXiv:2602.12281 [cs.RO]},
}

@misc{xiao_self-improving_2025,
	title = {Self-{Improving} {Vision}-{Language}-{Action} {Models} with {Data} {Generation} via {Residual} {RL}},
	url = {http://arxiv.org/abs/2511.00091},
	doi = {10.48550/arXiv.2511.00091},
	language = {en},
	urldate = {2026-09-25},
	publisher = {arXiv},
	author = {Xiao, Wenli and Lin, Haotian and Peng, Andy and Xue, Haoru and He, Tairan and Xie, Yuqi and Hu, Fengyuan and Wu, Jimmy and Luo, Zhengyi and Fan, Linxi "Jim" and Shi, Guanya and Zhu, Yuke},
	month = oct,
	year = {2025},
	note = {arXiv:2511.00091 [cs.CV]},
}

@misc{nguyen_self-improving_2026,
	title = {Self-{Improving} {VLA} {Policies}: {Selected} {Diffusion} {Noise} for {Spurious}-{Robust} {Action} {Smoothing}},
	shorttitle = {Self-{Improving} {VLA} {Policies}},
	url = {http://arxiv.org/abs/2606.14084},
	doi = {10.48550/arXiv.2606.14084},
	language = {en},
	urldate = {2026-09-25},
	publisher = {arXiv},
	author = {Nguyen, Duc Minh and Dao, Bao-Ngoc and Luu, Tung M. and Nguyen, Binh Gia and Tong, Vinh and Liu, Anji and Duong, Vu N. and Le, Dung D. and Sonntag, Daniel and Le, Trung and Le, Ngan and Peter, Jan and Le, An Thai and Vu, Minh Nhat and Niepert, Mathias and Doan, Khoa D. and Nguyen, Duy M. H. and Ngo, Vien Anh},
	month = jun,
	year = {2026},
	note = {arXiv:2606.14084 [cs.RO]},
}

@misc{li_simplevla-rl_2025,
	title = {{SimpleVLA}-{RL}: {Scaling} {VLA} {Training} via {Reinforcement} {Learning}},
	shorttitle = {{SimpleVLA}-{RL}},
	url = {http://arxiv.org/abs/2509.09674},
	doi = {10.48550/arXiv.2509.09674},
	language = {en},
	urldate = {2026-09-25},
	publisher = {arXiv},
	author = {Li, Haozhan and Zuo, Yuxin and Yu, Jiale and Zhang, Yuhao and Yang, Zhaohui and Zhang, Kaiyan and Zhu, Xuekai and Zhang, Yuchen and Chen, Tianxing and Cui, Ganqu and Wang, Dehui and Luo, Dingxiang and Fan, Yuchen and Sun, Youbang and Zeng, Jia and Pang, Jiangmiao and Zhang, Shanghang and Wang, Yu and Mu, Yao and Zhou, Bowen and Ding, Ning},
	month = sep,
	year = {2025},
	note = {arXiv:2509.09674 [cs.RO]},
}

@misc{yang_steering_2025,
	title = {Steering {Vision}-{Language}-{Action} {Models} as {Anti}-{Exploration}: {A} {Test}-{Time} {Scaling} {Approach}},
	shorttitle = {Steering {Vision}-{Language}-{Action} {Models} as {Anti}-{Exploration}},
	url = {http://arxiv.org/abs/2512.02834},
	doi = {10.48550/arXiv.2512.02834},
	language = {en},
	urldate = {2026-09-25},
	publisher = {arXiv},
	author = {Yang, Siyuan and Zhang, Yang and He, Haoran and Pan, Ling and Li, Xiu and Bai, Chenjia and Li, Xuelong},
	month = dec,
	year = {2025},
	note = {arXiv:2512.02834 [cs.RO]},
}

@misc{wagenmaker_steering_2025,
	title = {Steering {Your} {Diffusion} {Policy} with {Latent} {Space} {Reinforcement} {Learning}},
	url = {http://arxiv.org/abs/2506.15799},
	doi = {10.48550/arXiv.2506.15799},
	language = {en},
	urldate = {2026-09-25},
	publisher = {arXiv},
	author = {Wagenmaker, Andrew and Nakamoto, Mitsuhiko and Zhang, Yunchu and Park, Seohong and Yagoub, Waleed and Nagabandi, Anusha and Gupta, Abhishek and Levine, Sergey},
	month = jun,
	year = {2025},
	note = {arXiv:2506.15799 [cs.RO]},
}

@misc{nakamoto_steering_2025,
	title = {Steering {Your} {Generalists}: {Improving} {Robotic} {Foundation} {Models} via {Value} {Guidance}},
	shorttitle = {Steering {Your} {Generalists}},
	url = {http://arxiv.org/abs/2410.13816},
	doi = {10.48550/arXiv.2410.13816},
	language = {en},
	urldate = {2026-09-25},
	publisher = {arXiv},
	author = {Nakamoto, Mitsuhiko and Mees, Oier and Kumar, Aviral and Levine, Sergey},
	month = feb,
	year = {2025},
	note = {arXiv:2410.13816 [cs.RO]},
}

@misc{zhao_test-time_2026,
	title = {Test-{Time} {Scaling} for {World} {Action} {Models} via {Zero}-{Shot} {Geometric} {Evaluation}},
	url = {http://arxiv.org/abs/2607.17454},
	doi = {10.48550/arXiv.2607.17454},
	language = {en},
	urldate = {2026-09-25},
	publisher = {arXiv},
	author = {Zhao, Zesen and Cho, Minkyoung and shen, Hui and Zheng, Boyuan and Gao, Kunxiao and Cao, Yulong and Mao, Z. Morley},
	month = jul,
	year = {2026},
	note = {arXiv:2607.17454 [cs.RO]},
}

@misc{chen_tgrpo_2025,
	title = {{TGRPO} :{Fine}-tuning {Vision}-{Language}-{Action} {Model} via {Trajectory}-wise {Group} {Relative} {Policy} {Optimization}},
	shorttitle = {{TGRPO}},
	url = {http://arxiv.org/abs/2506.08440},
	doi = {10.48550/arXiv.2506.08440},
	language = {en},
	urldate = {2026-09-25},
	publisher = {arXiv},
	author = {Chen, Zengjue and Niu, Runliang and Kong, He and Wang, Qi and Xing, Qianli and Fan, Zipei},
	month = sep,
	year = {2025},
	note = {arXiv:2506.08440 [cs.RO]},
}

@misc{jang_verifier-free_2026,
	title = {Verifier-free {Test}-{Time} {Sampling} for {Vision}-{Language}-{Action} {Models}},
	url = {http://arxiv.org/abs/2510.05681},
	doi = {10.48550/arXiv.2510.05681},
	language = {en},
	urldate = {2026-09-25},
	publisher = {arXiv},
	author = {Jang, Suhyeok and Kim, Dongyoung and Kim, Changyeon and Kim, Youngsuk and Shin, Jinwoo},
	month = jul,
	year = {2026},
	note = {arXiv:2510.05681 [cs.RO]},
}

@misc{zhao_verispace_2026,
	title = {{VeriSpace}: {Spatially} {Grounded} {Action} {Verification} for {Vision}-{Language}-{Action} {Models}},
	shorttitle = {{VeriSpace}},
	url = {http://arxiv.org/abs/2606.10568},
	doi = {10.48550/arXiv.2606.10568},
	language = {en},
	urldate = {2026-09-25},
	publisher = {arXiv},
	author = {Zhao, Guiyu and Guo, Longteng and Zhu, Junyou and Fu, Jun and Mei, Yanghong and Cao, Bin and Jiang, Jie and He, Xingjian and Liu, Jing},
	month = jun,
	year = {2026},
	note = {arXiv:2606.10568 [cs.RO]},
}

@misc{wang_vla_2025,
	title = {{VLA} {Model} {Post}-{Training} via {Action}-{Chunked} {PPO} and {Self} {Behavior} {Cloning}},
	url = {http://arxiv.org/abs/2509.25718},
	doi = {10.48550/arXiv.2509.25718},
	language = {en},
	urldate = {2026-09-25},
	publisher = {arXiv},
	author = {Wang, Si-Cheng and Xiang, Tian-Yu and Zhou, Xiao-Hu and Gui, Mei-Jiang and Xie, Xiao-Liang and Liu, Shi-Qi and Wang, Shuang-Yi and Jin, Ao-Qun and Hou, Zeng-Guang},
	month = sep,
	year = {2025},
	note = {arXiv:2509.25718 [cs.RO]},
}

@misc{guo_vla-reasoner_2026,
	title = {{VLA}-{Reasoner}: {Empowering} {Vision}-{Language}-{Action} {Models} with {Reasoning} via {Online} {Monte} {Carlo} {Tree} {Search}},
	shorttitle = {{VLA}-{Reasoner}},
	url = {http://arxiv.org/abs/2509.22643},
	doi = {10.48550/arXiv.2509.22643},
	language = {en},
	urldate = {2026-09-25},
	publisher = {arXiv},
	author = {Guo, Wenkai and Lu, Guanxing and Deng, Haoyuan and Wu, Zhenyu and Tang, Yansong and Wang, Ziwei},
	month = mar,
	year = {2026},
	note = {arXiv:2509.22643 [cs.RO]},
}

@misc{lu_vla-rl_2025,
	title = {{VLA}-{RL}: {Towards} {Masterful} and {General} {Robotic} {Manipulation} with {Scalable} {Reinforcement} {Learning}},
	shorttitle = {{VLA}-{RL}},
	url = {http://arxiv.org/abs/2505.18719},
	doi = {10.48550/arXiv.2505.18719},
	language = {en},
	urldate = {2026-09-25},
	publisher = {arXiv},
	author = {Lu, Guanxing and Guo, Wenkai and Zhang, Chubin and Zhou, Yuheng and Jiang, Haonan and Gao, Zifeng and Tang, Yansong and Wang, Ziwei},
	month = may,
	year = {2025},
	note = {arXiv:2505.18719 [cs.RO]},
}

@misc{liu_what_2026,
	title = {What {Can} {RL} {Bring} to {VLA} {Generalization}? {An} {Empirical} {Study}},
	shorttitle = {What {Can} {RL} {Bring} to {VLA} {Generalization}?},
	url = {http://arxiv.org/abs/2505.19789},
	doi = {10.48550/arXiv.2505.19789},
	language = {en},
	urldate = {2026-09-25},
	publisher = {arXiv},
	author = {Liu, Jijia and Gao, Feng and Wei, Bingwen and Chen, Xinlei and Liao, Qingmin and Wu, Yi and Yu, Chao and Wang, Yu},
	month = jan,
	year = {2026},
	note = {arXiv:2505.19789 [cs.LG]},
}

@misc{cai__0-vla_2026,
	title = {$\tau_0$-{VLA}: a {Hierarchical} {Robot} {Foundation} {Model} with {World}-{Model}-{Guided} {Test}-{Time} {Computation}},
	shorttitle = {\$τ\_0\$-{VLA}},
	url = {http://arxiv.org/abs/2608.16885},
	doi = {10.48550/arXiv.2608.16885},
	language = {en},
	urldate = {2026-09-25},
	publisher = {arXiv},
	author = {Cai, Xiaowei and Cai, Yunuo and Chen, Bingao and Chen, Jingxiao and Chen, Zhi and Feng, Siyuan and Hou, Tengyu and Huang, Jingshun and Jiang, Han and Ju, Runkun and Li, Dong and Li, Mingxiang and Li, Shaowei and Li, Xinchen and Li, Yifan and Liu, Yi and Liu, Zhongyuan and Luo, Jianlan and Miao, Junwen and Ni, Ruiqi and Nie, Buqing and Pan, Mingjie and Ren, Xinlin and Song, Jianheng and Wang, Jiaxu and Wang, Peiqi and Wang, Sen and Wang, Xiaoyan and Wei, Dafeng and Wu, Dongming and Xie, Pengwei and Yang, Pu and Ye, Hangjian and Yue, Xiangyu and Zhang, Jinyu and Zhang, Qinglin and Zhao, Xueyong and Zhou, Pengfei and Zhou, Yue},
	month = aug,
	year = {2026},
	note = {arXiv:2608.16885 [cs.RO]},
}

@misc{zhai_vision-language-action-critic_2025,
	title = {A {Vision}-{Language}-{Action}-{Critic} {Model} for {Robotic} {Real}-{World} {Reinforcement} {Learning}},
	url = {http://arxiv.org/abs/2509.15937},
	doi = {10.48550/arXiv.2509.15937},
	language = {en},
	urldate = {2026-09-25},
	publisher = {arXiv},
	author = {Zhai, Shaopeng and Zhang, Qi and Zhang, Tianyi and Huang, Fuxian and Zhang, Haoran and Zhou, Ming and Zhang, Shengzhe and Liu, Litao and Lin, Sixu and Pang, Jiangmiao},
	month = sep,
	year = {2025},
	note = {arXiv:2509.15937 [cs.RO]},
}

@misc{giridhar_beyond_2026,
	title = {Beyond {Imitation}: {Self}-{Improving} {Robot} {Policies} via {Off}-{Policy} {Q}-{Planning}},
	shorttitle = {Beyond {Imitation}},
	url = {http://arxiv.org/abs/2608.21204},
	doi = {10.48550/arXiv.2608.21204},
	language = {en},
	urldate = {2026-09-25},
	publisher = {arXiv},
	author = {Giridhar, Varun and Khandelwal, Anant and Collins, Jeremy A. and Georgiev, Ignat and Garg, Animesh},
	month = aug,
	year = {2026},
	note = {arXiv:2608.21204 [cs.RO]},
}

@misc{singh_bokbo_2026,
	title = {{BOKBO} ({Best} of {K} {Bad} {Options}): {Calibrated} {Abstention} for {VLA} {Policies}},
	shorttitle = {{BOKBO} ({Best} of {K} {Bad} {Options})},
	url = {http://arxiv.org/abs/2605.30660},
	doi = {10.48550/arXiv.2605.30660},
	language = {en},
	urldate = {2026-09-25},
	publisher = {arXiv},
	author = {Singh, Anya and Happi, Cabrel and Relan, Jai and Nair, Varun and Baradwaj, Vidyut},
	month = may,
	year = {2026},
	note = {arXiv:2605.30660 [cs.LG]},
}

@misc{chen_conrft_2025,
	title = {{ConRFT}: {A} {Reinforced} {Fine}-tuning {Method} for {VLA} {Models} via {Consistency} {Policy}},
	shorttitle = {{ConRFT}},
	url = {http://arxiv.org/abs/2502.05450},
	doi = {10.48550/arXiv.2502.05450},
	language = {en},
	urldate = {2026-09-25},
	publisher = {arXiv},
	author = {Chen, Yuhui and Tian, Shuai and Liu, Shugao and Zhou, Yingting and Li, Haoran and Zhao, Dongbin},
	month = apr,
	year = {2025},
	note = {arXiv:2502.05450 [cs.RO]},
}

@misc{kim_deas_2025,
	title = {{DEAS}: {DEtached} value learning with {Action} {Sequence} for {Scalable} {Offline} {RL}},
	shorttitle = {{DEAS}},
	url = {http://arxiv.org/abs/2510.07730},
	doi = {10.48550/arXiv.2510.07730},
	language = {en},
	urldate = {2026-09-25},
	publisher = {arXiv},
	author = {Kim, Changyeon and Lee, Haeone and Seo, Younggyo and Lee, Kimin and Zhu, Yuke},
	month = oct,
	year = {2025},
	note = {arXiv:2510.07730 [cs.LG]},
}

@misc{chi_diffusion_2024,
	title = {Diffusion {Policy}: {Visuomotor} {Policy} {Learning} via {Action} {Diffusion}},
	shorttitle = {Diffusion {Policy}},
	url = {http://arxiv.org/abs/2303.04137},
	doi = {10.48550/arXiv.2303.04137},
	language = {en},
	urldate = {2026-09-25},
	publisher = {arXiv},
	author = {Chi, Cheng and Xu, Zhenjia and Feng, Siyuan and Cousineau, Eric and Du, Yilun and Burchfiel, Benjamin and Tedrake, Russ and Song, Shuran},
	month = mar,
	year = {2024},
	note = {arXiv:2303.04137 [cs.RO]},
}

@misc{bai_evolve-vla_2025,
	title = {{EVOLVE}-{VLA}: {Test}-{Time} {Training} from {Environment} {Feedback} for {Vision}-{Language}-{Action} {Models}},
	shorttitle = {{EVOLVE}-{VLA}},
	url = {http://arxiv.org/abs/2512.14666},
	doi = {10.48550/arXiv.2512.14666},
	language = {en},
	urldate = {2026-09-25},
	publisher = {arXiv},
	author = {Bai, Zechen and Gao, Chen and Shou, Mike Zheng},
	month = dec,
	year = {2025},
	note = {arXiv:2512.14666 [cs.RO]},
}

@misc{dong_expo_2026,
	title = {{EXPO}: {Stable} {Reinforcement} {Learning} with {Expressive} {Policies}},
	shorttitle = {{EXPO}},
	url = {http://arxiv.org/abs/2507.07986},
	doi = {10.48550/arXiv.2507.07986},
	language = {en},
	urldate = {2026-09-25},
	publisher = {arXiv},
	author = {Dong, Perry and Li, Qiyang and Sadigh, Dorsa and Finn, Chelsea},
	month = apr,
	year = {2026},
	note = {arXiv:2507.07986 [cs.LG]},
}

@misc{dong_expo-ft_2026,
	title = {{EXPO}-{FT}: {Sample}-{Efficient} {Reinforcement} {Learning} {Finetuning} for {Vision}-{Language}-{Action} {Models}},
	shorttitle = {{EXPO}-{FT}},
	url = {http://arxiv.org/abs/2605.25477},
	doi = {10.48550/arXiv.2605.25477},
	language = {en},
	urldate = {2026-09-25},
	publisher = {arXiv},
	author = {Dong, Perry and Hung, Kuo-Han and Gao, Tian and Sadigh, Dorsa and Finn, Chelsea},
	month = aug,
	year = {2026},
	note = {arXiv:2605.25477 [cs.RO]},
}

@misc{lipman_flow_2023,
	title = {Flow {Matching} for {Generative} {Modeling}},
	url = {http://arxiv.org/abs/2210.02747},
	doi = {10.48550/arXiv.2210.02747},
	language = {en},
	urldate = {2026-09-25},
	publisher = {arXiv},
	author = {Lipman, Yaron and Chen, Ricky T. Q. and Ben-Hamu, Heli and Nickel, Maximilian and Le, Matt},
	month = feb,
	year = {2023},
	note = {arXiv:2210.02747 [cs.LG]},
}

@misc{yang_guided_2026,
	title = {Guided {Action} {Flow}: {Q}-{Guided} {Inference} for {Flow}-{Matching} {Vision}-{Language}-{Action} {Policies}},
	shorttitle = {Guided {Action} {Flow}},
	url = {http://arxiv.org/abs/2607.02092},
	doi = {10.48550/arXiv.2607.02092},
	language = {en},
	urldate = {2026-09-25},
	publisher = {arXiv},
	author = {Yang, Liuhaichen and Jiang, Zhuang and Sheng, Chenchao and Tang, Zezhi},
	month = aug,
	year = {2026},
	note = {arXiv:2607.02092 [cs.RO]},
}

@misc{wang_insight_2026,
	title = {{InSight}: {Self}-{Guided} {Skill} {Acquisition} via {Steerable} {VLAs}},
	shorttitle = {{InSight}},
	url = {http://arxiv.org/abs/2606.24884},
	doi = {10.48550/arXiv.2606.24884},
	language = {en},
	urldate = {2026-09-25},
	publisher = {arXiv},
	author = {Wang, Maggie and Osterberg, Lars and Tian, Stephen and Shorinwa, Ola and Bowkett, Joseph and Pastor, Daniel and Wu, Jiajun and Schwager, Mac},
	month = sep,
	year = {2026},
	note = {arXiv:2606.24884 [cs.RO]},
}

@misc{wu_large_2026,
	title = {Large {Reward} {Models}: {Generalizable} {Online} {Robot} {Reward} {Generation} with {Vision}-{Language} {Models}},
	shorttitle = {Large {Reward} {Models}},
	url = {http://arxiv.org/abs/2603.16065},
	doi = {10.48550/arXiv.2603.16065},
	language = {en},
	urldate = {2026-09-25},
	publisher = {arXiv},
	author = {Wu, Yanru and Yuan, Weiduo and Qi, Ang and Guizilini, Vitor and Mao, Jiageng and Wang, Yue},
	month = mar,
	year = {2026},
	note = {arXiv:2603.16065 [cs.RO]},
}

@misc{liu_--fly_2026,
	title = {On-the-{Fly} {VLA} {Adaptation} via {Test}-{Time} {Reinforcement} {Learning}},
	url = {http://arxiv.org/abs/2601.06748},
	doi = {10.48550/arXiv.2601.06748},
	language = {en},
	urldate = {2026-09-25},
	publisher = {arXiv},
	author = {Liu, Changyu and Liu, Yiyang and Wang, Taowen and Zhuang, Qiao and Liang, James Chenhao and Yang, Wenhao and Xu, Renjing and Wang, Qifan and Liu, Dongfang and Han, Cheng},
	month = apr,
	year = {2026},
	note = {arXiv:2601.06748 [cs.RO]},
}

@misc{julg_refined_2025,
	title = {Refined {Policy} {Distillation}: {From} {VLA} {Generalists} to {RL} {Experts}},
	shorttitle = {Refined {Policy} {Distillation}},
	url = {http://arxiv.org/abs/2503.05833},
	doi = {10.48550/arXiv.2503.05833},
	language = {en},
	urldate = {2026-09-25},
	publisher = {arXiv},
	author = {Jülg, Tobias and Burgard, Wolfram and Walter, Florian},
	month = aug,
	year = {2025},
	note = {arXiv:2503.05833 [cs.RO]},
}

@misc{li_reflection-based_2026,
	title = {Reflection-{Based} {Task} {Adaptation} for {Self}-{Improving} {VLA}},
	url = {http://arxiv.org/abs/2510.12710},
	doi = {10.48550/arXiv.2510.12710},
	language = {en},
	urldate = {2026-09-25},
	publisher = {arXiv},
	author = {Li, Baicheng and Wu, Dong and Yan, Zike and Liu, Xinchen and Li, Lusong and Zeng, Zecui and Zha, Hongbin},
	month = apr,
	year = {2026},
	note = {arXiv:2510.12710 [cs.RO]},
}

@misc{yang_rise_2026,
	title = {{RISE}: {Self}-{Improving} {Robot} {Policy} with {Compositional} {World} {Model}},
	shorttitle = {{RISE}},
	url = {http://arxiv.org/abs/2602.11075},
	doi = {10.48550/arXiv.2602.11075},
	language = {en},
	urldate = {2026-09-25},
	publisher = {arXiv},
	author = {Yang, Jiazhi and Lin, Kunyang and Li, Jinwei and Zhang, Wencong and Lin, Tianwei and Wu, Longyan and Su, Zhizhong and Zhao, Hao and Zhang, Ya-Qin and Chen, Li and Luo, Ping and Yue, Xiangyu and Li, Hongyang},
	month = apr,
	year = {2026},
	note = {arXiv:2602.11075 [cs.RO]},
}

@misc{tan_rl2-vla_2026,
	title = {{RL}\${\textasciicircum}2\$-{VLA}: {Adaptive} {RL} {Latent} {Compositional} {Steering} with {Test}-{Time} {Scaling} for {Vision}-{Language}-{Action} {Models}},
	shorttitle = {{RL}\${\textasciicircum}2\$-{VLA}},
	url = {http://arxiv.org/abs/2607.26991},
	doi = {10.48550/arXiv.2607.26991},
	language = {en},
	urldate = {2026-09-25},
	publisher = {arXiv},
	author = {Tan, Derek Ming Siang and Shailesh, Shailesh and Iyer, Srikrishna and Teo, William Wei Jie and Ju, Yuanliang and Gu, Qiao and Sartoretti, Guillaume},
	month = jul,
	year = {2026},
	note = {arXiv:2607.26991 [cs.RO]},
}

@misc{kwok_robomonkey_2025,
	title = {{RoboMonkey}: {Scaling} {Test}-{Time} {Sampling} and {Verification} for {Vision}-{Language}-{Action} {Models}},
	shorttitle = {{RoboMonkey}},
	url = {http://arxiv.org/abs/2506.17811},
	doi = {10.48550/arXiv.2506.17811},
	language = {en},
	urldate = {2026-09-25},
	publisher = {arXiv},
	author = {Kwok, Jacky and Agia, Christopher and Sinha, Rohan and Foutter, Matt and Li, Shulu and Stoica, Ion and Mirhoseini, Azalia and Pavone, Marco},
	month = jul,
	year = {2025},
	note = {arXiv:2506.17811 [cs.RO]},
}

@misc{dai_rover_2025,
	title = {{RoVer}: {Robot} {Reward} {Model} as {Test}-{Time} {Verifier} for {Vision}-{Language}-{Action} {Model}},
	shorttitle = {{RoVer}},
	url = {http://arxiv.org/abs/2510.10975},
	doi = {10.48550/arXiv.2510.10975},
	language = {en},
	urldate = {2026-09-25},
	publisher = {arXiv},
	author = {Dai, Mingtong and Liu, Lingbo and Bai, Yongjie and Liu, Yang and Wang, Zhouxia and SU, Rui and Chen, Chunjie and Lin, Liang and Wu, Xinyu},
	month = oct,
	year = {2025},
	note = {arXiv:2510.10975 [cs.RO]},
}

@misc{shukor_smolvla_2025,
	title = {{SmolVLA}: {A} {Vision}-{Language}-{Action} {Model} for {Affordable} and {Efficient} {Robotics}},
	shorttitle = {{SmolVLA}},
	url = {http://arxiv.org/abs/2506.01844},
	doi = {10.48550/arXiv.2506.01844},
	language = {en},
	urldate = {2026-09-25},
	publisher = {arXiv},
	author = {Shukor, Mustafa and Aubakirova, Dana and Capuano, Francesco and Kooijmans, Pepijn and Palma, Steven and Zouitine, Adil and Aractingi, Michel and Pascal, Caroline and Russi, Martino and Marafioti, Andres and Alibert, Simon and Cord, Matthieu and Wolf, Thomas and Cadene, Remi},
	month = jun,
	year = {2025},
	note = {arXiv:2506.01844 [cs.LG]},
}

@misc{liu_libero_2023,
	title = {{LIBERO}: {Benchmarking} {Knowledge} {Transfer} for {Lifelong} {Robot} {Learning}},
	shorttitle = {{LIBERO}},
	url = {http://arxiv.org/abs/2306.03310},
	doi = {10.48550/arXiv.2306.03310},
	language = {en},
	urldate = {2026-09-25},
	publisher = {arXiv},
	author = {Liu, Bo and Zhu, Yifeng and Gao, Chongkai and Feng, Yihao and Liu, Qiang and Zhu, Yuke and Stone, Peter},
	month = oct,
	year = {2023},
	note = {arXiv:2306.03310 [cs.AI]},
}

@misc{fei_libero-plus_2025,
	title = {{LIBERO}-{Plus}: {In}-depth {Robustness} {Analysis} of {Vision}-{Language}-{Action} {Models}},
	shorttitle = {{LIBERO}-{Plus}},
	url = {http://arxiv.org/abs/2510.13626},
	doi = {10.48550/arXiv.2510.13626},
	language = {en},
	urldate = {2026-09-25},
	publisher = {arXiv},
	author = {Fei, Senyu and Wang, Siyin and Shi, Junhao and Dai, Zihao and Cai, Jikun and Qian, Pengfang and Ji, Li and He, Xinzhe and Zhang, Shiduo and Fei, Zhaoye and Fu, Jinlan and Gong, Jingjing and Qiu, Xipeng},
	month = dec,
	year = {2025},
	note = {arXiv:2510.13626 [cs.RO]},
}

@misc{hu_lora_2021,
	title = {{LoRA}: {Low}-{Rank} {Adaptation} of {Large} {Language} {Models}},
	shorttitle = {{LoRA}},
	url = {http://arxiv.org/abs/2106.09685},
	doi = {10.48550/arXiv.2106.09685},
	language = {en},
	urldate = {2026-09-25},
	publisher = {arXiv},
	author = {Hu, Edward J. and Shen, Yelong and Wallis, Phillip and Allen-Zhu, Zeyuan and Li, Yuanzhi and Wang, Shean and Wang, Lu and Chen, Weizhu},
	month = oct,
	year = {2021},
	note = {arXiv:2106.09685 [cs.CL]},
}

@misc{intelligence__05_2025,
	title = {\${\(\pi_{0.5}\)}\$: a {Vision}-{Language}-{Action} {Model} with {Open}-{World} {Generalization}},
	shorttitle = {\$π\_\{0.5\}\$},
	url = {http://arxiv.org/abs/2504.16054},
	doi = {10.48550/arXiv.2504.16054},
	language = {en},
	urldate = {2026-09-25},
	publisher = {arXiv},
	author = {Intelligence, Physical and Black, Kevin and Brown, Noah and Darpinian, James and Dhabalia, Karan and Driess, Danny and Esmail, Adnan and Equi, Michael and Finn, Chelsea and Fusai, Niccolo and Galliker, Manuel Y. and Ghosh, Dibya and Groom, Lachy and Hausman, Karol and Ichter, Brian and Jakubczak, Szymon and Jones, Tim and Ke, Liyiming and LeBlanc, Devin and Levine, Sergey and Li-Bell, Adrian and Mothukuri, Mohith and Nair, Suraj and Pertsch, Karl and Ren, Allen Z. and Shi, Lucy Xiaoyang and Smith, Laura and Springenberg, Jost Tobias and Stachowicz, Kyle and Tanner, James and Vuong, Quan and Walke, Homer and Walling, Anna and Wang, Haohuan and Yu, Lili and Zhilinsky, Ury},
	month = apr,
	year = {2025},
	note = {arXiv:2504.16054 [cs.LG]},
}

@article{bradley_terry,
 ISSN = {00063444, 14643510},
 URL = {http://www.jstor.org/stable/2332730},
 author = {Ralph Allan Bradley},
 journal = {Biometrika},
 number = {3/4},
 pages = {502--537},
 publisher = {[Oxford University Press, Biometrika Trust]},
 title = {Rank Analysis of Incomplete Block Designs: II. Additional Tables for the Method of Paired Comparisons},
 urldate = {2026-09-25},
 volume = {41},
 year = {1954}
}

@misc{mark2024policyagnostic,
      title={Policy Agnostic RL: Offline RL and Online RL Fine-Tuning of Any Class and Backbone}, 
      author={Max Sobol Mark and Tian Gao and Georgia Gabriela Sampaio and Mohan Kumar Srirama and Archit Sharma and Chelsea Finn and Aviral Kumar},
      year={2024},
      eprint={2412.06685},
      archivePrefix={arXiv},
      primaryClass={cs.LG},
      url={https://arxiv.org/abs/2412.06685}, 
}

@misc{xie2026look,
      title={Look Before You Leap: Distilling Tree Search into Action Evaluation for Frozen VLA Models}, 
      author={Xinyi Xie and Zican Hu and Zhanyu Liu and Yicheng Dong and Wenhao Wu and Zhenhong Sun and Haoran Li and Chunlin Chen and Zhi Wang and Pichao Wang},
      year={2026},
      eprint={2607.03751},
      archivePrefix={arXiv},
      primaryClass={cs.RO},
      url={https://arxiv.org/abs/2607.03751}, 
}

@misc{wang2026qvgm,
      title={Q-VGM: Q-Guided Value-Gradient Matching for Offline-to-Online RL of Flow-Matching VLA Policies}, 
      author={Ziqian Wang and Rui Zhang and Yitian Liu and Xingjian Mao and Minqian Wang and Yao Mu},
      year={2026},
      eprint={2606.08015},
      archivePrefix={arXiv},
      primaryClass={cs.RO},
      url={https://arxiv.org/abs/2606.08015}, 
}

@misc{anthony2017thinking,
      title={Thinking Fast and Slow with Deep Learning and Tree Search}, 
      author={Thomas Anthony and Zheng Tian and David Barber},
      year={2017},
      eprint={1705.08439},
      archivePrefix={arXiv},
      primaryClass={cs.AI},
      url={https://arxiv.org/abs/1705.08439}, 
}
